\pdfoutput=1
\documentclass[11pt]{article}

\usepackage{acl}

\usepackage{times}
\usepackage{latexsym}

\usepackage[T1]{fontenc}
\usepackage[utf8]{inputenc}

\usepackage{microtype}

\usepackage{soul}

\newcommand{\draftonly}[1]{#1}
\renewcommand{\draftonly}[1]{}

\newcommand{\esnli}{\textsc{E-SNLI}\xspace}
\newcommand{\cose}{\textsc{CoS-E}\xspace}
\newcommand{\ecqa}{\textsc{ECQA}\xspace}
\newcommand{\sbic}{\textsc{SBIC}\xspace}
\newcommand{\comve}{\textsc{ComVE}\xspace}
\newcommand{\unifiedqa}{\textsc{UnifiedQA}\xspace}
\newcommand{\unifew}{\textsc{UniFew}\xspace}

\newcommand{\infilling}{\textsc{\colorbox{red!10}{Infilling}}\xspace}
\newcommand{\simtfive}{\textsc{\colorbox{orange!10}{$\approx$\textsc{T5}}}\xspace}
\newcommand{\unifiedqaprompt}{\textsc{\colorbox{blue!10}{\textsc{QA}$_{\textnormal{\scriptsize SIMPLE}}$}}\xspace}
\newcommand{\squadprompt}{\textsc{\colorbox{green!10}{\textsc{SQuAD$_{\textnormal{\scriptsize T5}}$}}}\xspace}

\newcommand\sect[1]{\S\ref{#1}}
\newcommand{\benchmark}{\textsc{FEB}\xspace}

\usepackage{paralist}
\usepackage{multirow}
\usepackage{graphicx}
\usepackage{booktabs}
\usepackage{colortbl}
\usepackage{makecell}
\usepackage{xspace}
\usepackage[many]{tcolorbox}
\usepackage{inconsolata}
\usepackage{subcaption}
\usepackage{booktabs}

\title{Few-Shot Self-Rationalization with Natural Language Prompts}

\author{
	Ana Marasovi\'{c}\thanks{\enspace Equal contributions.}  \quad
	Iz Beltagy\footnotemark[1]  \quad
	Doug Downey  \quad 
	Matthew E.\ Peters  \\\\
	Allen Institute for AI, Seattle, WA, USA \\
	{\tt \{anam,beltagy,dougd,matthewp\}@allenai.org}
}

\begin{document}
\maketitle
\begin{abstract}
Self-rationalization models that predict task labels and generate free-text elaborations for their predictions could enable more intuitive interaction with NLP systems. 
These models are, however, currently trained with a large amount of human-written free-text explanations for each task which hinders their broader usage. 
We propose to study a more realistic setting of self-rationalization using few training examples. 
We present \benchmark---a standardized collection of four existing English-language datasets and associated metrics. 
We identify the right prompting approach by extensively exploring natural language prompts on \benchmark. 
Then, by using this prompt and scaling the model size, we demonstrate that making progress on few-shot self-rationalization is possible. 
We show there is still ample room for improvement in this task: the average plausibility of generated explanations assessed by human annotators is at most 51\% (with GPT-3), while plausibility of human explanations is 76\%. 
We hope that \benchmark and our proposed approach will spur the community to take on the few-shot self-rationalization challenge. 

\end{abstract}

\section{Introduction}

Models constrained to be more understandable to people are easier to troubleshoot and more useful in practice \cite{Rudin2021InterpretableML}. %
For instance, constraining a model that answers the question ``Which linguist invented the lightbulb?'' with ``none'' to also provide the reason---``Thomas Edison is the inventor of the lightbulb and he was not a linguist''---makes the model easier to control and interact with   \cite{Kim2021WhichLI}. %
Models that jointly predict task labels and generate \emph{free-text explanations} for their predictions (as in the previous example) are known as \emph{self-rationalization models} \cite{wiegreffe-etal-2021-measuring}. %
Their explanations are arguably more faithful and stable than post-hoc explanations since they are intrinsic to the model \cite{melis2018towards}. %
The free-text format is essential for explaining tasks requiring reasoning about unstated knowledge such as commonsense \cite{marasovic-etal-2020-natural}, and it makes explanations more intuitive to people compared to highlights of individual words \cite{camburu2018snli}. %   
Despite these benefits, self-rationalization models are not widely used, in part because their training currently requires an abundance of human-authored explanations for each task \cite{narang2020wt5}. %
A possible solution is few-shot learning, which has shown promising results in recent years. %
To help the research community begin tackling self-rationalization with only a few examples, we present (i) \benchmark---a standardized collection of four existing English-language datasets and associated metrics, and (ii) the first approach for the task established through an extensive evaluation of natural language prompts.\footnote{\textbf{F}ew \textbf{E}xplanations \textbf{B}enchmark (\textbf{FEB})} 

One approach to few-shot learning is \emph{prompt-based finetuning} with \emph{natural language prompts}. %
Such prompts are produced by formatting finetuning instances using a format similar to that used in pretraining, based on the idea that finetuning examples that look similar to pretraining ones will be more informative in the fewshot setting. %
A few prompts are then used for finetuning. %
In this paper, we explore whether prompt-based finetuning can be extended to induce few-shot self-rationalization behavior in addition to few-shot prediction. %
To measure our progress, we first introduce \benchmark as a benchmark dataset consisting of human authored free-text explanations across four distinct end tasks including natural language inference and commonsense tasks (\sect{sec:benchmark}). %
Since finding appropriate prompts is often challenging \cite{gao-etal-2021-making}, we then extensively explore  natural language prompts for few-shot self-rationalization.
In our experiments, we fine-tune the T5 and \unifiedqa pretrained encoder-decoder transformers \cite{raffel2019exploring,khashabi-etal-2020-unifiedqa}, and show that versatile question-answering prompts (defined in \sect{sec:qa_prompts}) outperform prompts based on span infilling by 8.7 accuracy points, as well as prompts designed by following the most similar T5's supervised pretraining task by 3.2. 

 We then study the impact of model size on few-shot self-rationalization to investigate whether the quality of generated explanations scales with the size as good as the accuracy of predicting task labels. To this end, we also evaluate \textsc{GPT-3}'s \cite{brown2020gpt3} self-rationalization behavior. 
 Our experiments show that explanation plausibility scored by human annotators (which range from 0--100) and end-task accuracy improve with increasing model size. % 
 Specifically, the difference in plausibility scores between the \textsc{base} and 3B model ranges from [6.2, 24.8] (on average 14.8). %
 The average plausibility across datasets is 43.4 (\unifiedqa-3B) and 50.6 (\textsc{GPT-3}). %
While encouraging, our results show that there is still a large gap between model and human performance (25.7 for \textsc{GPT-3}), and we hope this work will help enable the research community to take on the few-shot self-rationalization challenge. 

Our code for producing data splits, prompt construction, model training/evaluation, and human evaluation templates are publicly available.\footnote{\url{https://github.com/allenai/feb}}

\section{\benchmark Benchmark}
\label{sec:benchmark} 

There has been an explosion of interest in generating free-text explanations and in few-shot learning in the last 1--2 years. %
However, appropriate datasets and metrics for few-shot self-rationalization have not yet been established. %
We thus introduce the \benchmark benchmark---a suite of existing English-language datasets with human-authored free-text explanations and associated metrics for few-shot self-rationalization. %
We expect that \benchmark will simplify future model comparison and lower barriers to entry for those interested in working on this task. %

\paragraph{Datasets in \benchmark} To identify available datasets suitable for few-shot self-rationalization, we start with a recent overview of datasets with free-text explanations \cite{wiegreffe-marasovic-2021-review} and filter them according to the following criteria: 
(i) the input is textual, (ii) the explanation consists of one sentence or 2--3 simple sentences, (iii) the task has a fixed set of possible labels, (iv) the explanation is human-authored, and (v) the dataset has at least 389 instances. 
We use the second and third criteria to narrow the scope to easier self-rationalization since we expect that few-shot self-rationalization is very challenging. 
The last requirement is introduced to have 48 training and 350 evaluation examples. 

\begin{table}[t]
 \resizebox{\columnwidth}{!}{  
\begin{tabular}{lp{5.5cm}c}
\toprule
\multicolumn{2}{l}{\textbf{\benchmark Tasks}}   & \textbf{\# Shots} \\
\midrule
\textsc{\small E-SNLI} & Classify the entailment relation between two sequences & 16          \\
\arrayrulecolor{black!20}\midrule
\textsc{\small ECQA} & Select the correct answer to a given question from five answer choices & 48 \\
\arrayrulecolor{black!20}\midrule
\textsc{\small ComVE}  & Select one of two sequences as more nonsensical & 24 \\
\arrayrulecolor{black!20}\midrule
\textsc{\small SBIC} & Classify a post as offensive or not & 24      \\
\arrayrulecolor{black}\bottomrule
\end{tabular}
}
\caption{Tasks that we have included in \benchmark. The number of shots is the number of training instances \emph{per label}. Training sets for all classification tasks are balanced and contain 48 instances. Sources: \textsc{E-SNLI} \cite{camburu2018snli}, \textsc{ECQA} \cite{aggarwal-etal-2021-explanations},  \textsc{ComVE} \cite{wang-etal-2019-make}, \textsc{SBIC} \cite{sap-etal-2020-social}. 
}
\label{tab:tasks}
\end{table}

This gives us 5 datasets, 4 of which are included in \benchmark and overviewed in Table \ref{tab:tasks}. 
These datasets span 4 different tasks: natural language inference, multiple-choice commonsense QA, nonsensical sentence selection, and offensiveness classification. We exclude \cose \cite{rajani-etal-2019-explain} as it is too noisy to be useful for modeling and evaluating self-rationalization \cite{narang2020wt5}, but we do not support using \cose in the future, especially since \ecqa is introduced.\footnote{Since \cose is still actively used, we report  \cose results in Tables  \ref{tab:appendix_cos_various} and  \ref{tab:appendix_cose_scale} in Appendix.}

\ecqa contains not only justifications of the correct answer, but also justifications that refute the incorrect answer choices. %
We use only the former since they answer ``why is [input] assigned [label]?'', just as explanations in other datasets that we have included in \benchmark. %
The \sbic dataset contains annotations of frames representing the social biases that are implied in language. 
We format these frames as a self-rationalization task as follows. 
We allow only two labels: ``offensive'' and ``not offensive''. 
If a post is not offensive, we assign it the explanation: ``\emph{This post does not imply anything offensive}.'' 
A post can be offensive because it targets an individual or a demographic group. 
In the former cases, a post is assigned the explanation: ``\emph{This post is a personal attack}.'' 
Otherwise, we define a set of rules to transform \sbic annotations of which identity-based group is targeted and what stereotypes of this group are referenced or implied into a single, coherent sentence; e.g., \texttt{group}: \emph{women}, \texttt{stereotype}: \emph{can't drive} $\rightarrow$ ``\emph{This post is offensive because it implies that women can't drive}''. 

This is, to the best of our knowledge, the most comprehensive collection of self-rationalization tasks with textual inputs that could also be used even when working in a high-resource setting.

\paragraph{Automatic Evaluation} Evaluating self-rationalization (predicting task labels and generating explanations for the predicted labels) requires end-task evaluation and assessing the explanation plausibility. %
We use accuracy as our end-task evaluation metric. 
Explanation plausibility may be described as a subjective satisfaction with how a given explanation justifies a label/answer \cite{Yang2019EvaluatingEW}. 
\citet{Kayser2021eViLAD} present the largest currently available study on the correlation of 10 NLG metrics with human judgments of free-text explanation plausibility and report that BERTscore~\cite{Zhang2020BERTScoreET} is most correlated (although the correlation is still weak). 
Thus, we use BERTscore to evaluate the similarity between gold and generated explanations. %
Following \citeauthor{Kayser2021eViLAD}, we assign zero BERTscore to explanations of incorrectly predicted instances.\footnote{\citet{Kayser2021eViLAD}: ``An explanation is expected to be false when the answer is predicted incorrectly (as it is expected to justify a wrong answer).''} 

We follow recent recommendations for reliable few-shot evaluation \cite{Bragg2021FLEXUE}. %
Specifically, we fix hyperparameters (HPs) and use 60 random train-dev splits with 350 examples in each dev set. %
For classification tasks, the number of shots (examples per label) is chosen such that we construct a balanced training set of size 48.\footnote{In early studies, we found that 48 gives models that are at least slightly above the random baseline across all four tasks.} %
See Table \ref{tab:tasks} (col.\ 3) for exact values; for \ecqa we sample 48 training examples. %
For each model, we report the mean and standard error of 60 mean accuracy/BERTscore values calculated on 60 dev sets of 350 examples.\footnote{To calculate the standard error for accuracy/BERTscore we use $n=60$. The training (and likewise, dev) sets across splits can overlap, so this error reflects the variability expected in average scores when repeating our experiment with 60 new random splits of the same data sets.} 
Our HPs are reported in Table \ref{tab:hyperparameters} in Appendix. 
 
\paragraph{Human Evaluation} For our final models (\sect{sec:scaling}), we conduct a human evaluation of plausibility of generated explanations following prior work \cite{Kayser2021eViLAD, marasovic-etal-2020-natural}. %
For each model evaluation, \citet{Kayser2021eViLAD} take the first 300 dev examples that are correctly predicted by the model. %
This means that the dev set subsets used for human evaluation  differ across models that are evaluated. %
However, the overlap between the evaluation sets is maximized by fixing the order of dev instances and taking the first 300.

Prior work used a single train-dev split, while \benchmark has 60 train-dev splits. %
Multiple splits provides the opportunity to account for the variance caused by changing the random seed to produce a reliable estimate of plausibility of explanations produced with only a few examples. %
Therefore, we take the first 6 correctly predicted examples per train-dev split, i.e., 6*60=360 total instances. %
Moreover, for classification tasks, we propose to take the first 6/\texttt{\#}\textnormal{labels} correctly-predicted examples per label to have a balanced evaluation set. %

Following \citet{Kayser2021eViLAD}, we conduct the human evaluation in two steps:
\begin{compactitem}
\item \textbf{Step1}: Select the correct label/answer. 
\item \textbf{Step2}: Assess whether two explanations (gold and generated) justify the label/answer above. 
\end{compactitem}
The first step makes sure the annotators understood the task correctly and they are not able to submit their annotations if the answers are wrong.\footnote{We skipped this step for \ecqa because we could not teach crowdworkers sufficiently well to select the most likely answer out of multiple likely answer candidates in \ecqa.} 
Ground-truth explanations are evaluated to implicitly influence annotators with a gold reference point when they evaluate generated explanations, and to measure the quality of explanation datasets. 
To evaluate explanations, annotators are asked ``Does the explanation justify the answer?'' and given the options \texttt{\{``yes'',``weak yes'',``weak no'',``no''\}}. %
These options are mapped to plausibility scores of \texttt{\{1,$\frac{2}{3}$,$\frac{1}{3}$,0\}}, respectively. %
For each of the 360 examples, we calculate the mean plausibility score of 3 annotators and report the mean and the standard error of 360 mean scores. %
We also report the inter-annotator agreement calculated with Fleiss' kappa. % 
Finally, models are evaluated independently to avoid penalizing worse models in the presence of explanations generated by a better model.

\section{Prompting for Self-Rationalization} 
\label{sec:study_various_prompts}

We approach few-shot self-rationalization with prompt-based finetuning using natural language (NL) prompts. 
The key idea behind NL prompts is that a pretrained language model (LM) is already well-positioned to solve the end-task if we format finetuning end-task examples as similar as possible to the format used in the LM's pretraining. 
Following that principle, in this section, we describe our prompting approach with T5 \cite{raffel2019exploring} and comprehensively evaluate three distinct prompt types with \benchmark. 
Our results show that a unified question-answering (QA) prompt combined with a T5 variant that includes additional supervised multitask QA training (\unifiedqa; \citealp{khashabi-etal-2020-unifiedqa}) performs the best overall across tasks, when compared to three different alternative prompts as described below.

Self-rationalization models \cite{narang2020wt5, wiegreffe-etal-2021-measuring} are currently based on T5 for at least two reasons. 
First, T5 has been pretrained with many supervised tasks including classification and generation tasks, and self-rationalization involves both classification and generation. 
Second, T5 is one of the largest \emph{open-sourced} and widely studied pretrained models, and higher LM performance is correlated with larger model size \cite{Kaplan2020ScalingLF}. 
Thus, all of our experiments are based on T5 (and the \unifiedqa variant when evaluating prompts based on a QA format). 
In this section, all results are obtained with the base version of these models and in \sect{sec:scaling} we scale model size.   

When a LM is pretrained with masked language modeling  \cite{devlin-etal-2019-bert} only, an appropriate NL prompt is constructed by adding and infilling masked tokens \cite{jiang-etal-2020-know}. 
T5, however, has been pretrained with span infilling and a suite of supervised tasks whose instances were formatted in various ways. 
One of these supervised tasks includes \textsc{SQuAD 1.1} \cite{rajpurkar-etal-2016-squad} which allows us to experiment with prompts based on QA templates.   
As a result, we were able to design several different types of NL prompts for T5 consistent with different aspects of its pretraining: 
\begin{compactenum}
\item QA prompts (\squadprompt, \unifiedqaprompt).
\item span-filling prompts (\infilling),
\item prompts designed by following the formatting of the most similar T5's pretraining task (\simtfive; see Table \ref{tab:similar_tasks} in Appendix), 
\end{compactenum}
We illustrate these prompt types for \comve in Table \ref{tab:prompts_comve} in Appendix. 
The following sections describe these formats in detail and compare their performance using \benchmark.

\subsection{QA Prompts}
\label{sec:qa_prompts}

Formatting new instances as QA pairs has been shown to be useful for transfer learning from a QA model \cite{Gardner2019QuestionAI}. %
We first evaluate options for a versatile QA NL prompt for self-rationalization of tasks in \benchmark before comparing this approach with the other two prompt types (\infilling and \simtfive) in \sect{sec:main_comparison}. %
As alternative QA models, we investigate two models: T5 (which has been pretrained with QA supervision from \textsc{SQuAD 1.1}), and \unifiedqa (a T5 variant described in detail below). 
Since \unifiedqa was trained on a multitask mixture of many different QA datasets, these T5 variants allow us to examine the extent to which additional QA supervision can transfer to the few-shot self-rationalization setting.

Prior work \citep{Bragg2021FLEXUE} introduced \unifew, a model based on \unifiedqa, that is finetuned on a few task-specific instances posed as QA.
Despite its simplicity, \unifew achieves competitive few-shot learning performance with strong baselines for classification tasks. 
However, \citeauthor{Bragg2021FLEXUE}'s prompts do not cover all task types in \benchmark, and the question structure in their prompts is highly task-specific (see Appendix \ref{sec:related_work_formats}). 

Alternatively, we propose to design QA prompts with a simple principle in mind: Given a non-QA task, construct an equivalent QA task in the form of short ``Is...?''  or ``What is...?'' questions. Here, ``Is...?'' questions have yes/no answers (sometimes ``maybe''), and task labels verbatim are answers to ``What is...?'' questions (e.g., ``offensive'' and ``not offensive''). %
Given such question-answer pairs, we develop prompts following the formats proposed in \unifiedqa (see Appendix \ref{sec:appendix}) and prompt \unifiedqa. 
We denote these prompts as \unifiedqaprompt. 
For T5, we develop prompts following the \textsc{SQuAD} format for the T5's pretraining (\squadprompt; see Appendix \ref{sec:appendix}).  

There is another factor to consider. 
We need to decide whether to add \emph{tags}---a single description of each input element. 
Examples of tags are  ``\texttt{premise:}'' and ``\texttt{hypothesis:}'' before the first and second sentence in the \esnli input. 
Without these tags the task seems impossible to understand, but \unifiedqa has not been trained with any tags. 

The output always takes the form of ``[\emph{answer/label}] \texttt{because} [\emph{explanation}]''. See Table \ref{tab:prompts_comve} (Appendix) for  examples of our various QA prompts.

\paragraph{Results} We present the results of \unifiedqa with \unifiedqaprompt in Table \ref{tab:qa_prompts}, and due to space limits, T5's results with  \squadprompt prompts in Table \ref{tab:squad_qa_prompts} in Appendix. 

We observe that for \esnli and \comve it is crucial to add tags (``\texttt{premise:}''/``\texttt{hypothesis:}''; ``\texttt{choice1:}''/``\texttt{choice2:}'').\footnote{Performance on \comve with ``Is...?'' is close to random regardless of tags which suggests that this question form hinders the performance and tags cannot make a difference.}
This result is intuitive---it should be difficult to pick one of the two sentences, or classify a relation between them, if sentences are not marked. %
On the other hand, adding label choices is not beneficial and in some cases can even decrease the performance.  
When tags are included, we see that across all the tasks the ``What is...?'' question  performs the best.  
This also holds for T5 and \squadprompt prompts (see Table \ref{tab:squad_qa_prompts}). %
Finally, the prompt with the ``What is...?'' question and tags in the input outperforms \unifew for both tasks \unifew can be applied to. This result shows that this prompt is both  versatile and effective.

Finally, we compare the best performing prompts we get with \unifiedqa with \unifiedqaprompt and T5 wtih \squadprompt. 
See prompts ``\squadprompt $\times$ \textsc{What is...?} + \textsc{Tags}'' and ``\unifiedqaprompt $\times$ \textsc{What is...?} + \textsc{Tags}'' in Table \ref{tab:prompts_comve}. For \ecqa and \comve, we observe notable improvements from using \unifiedqa, and minor improvements for \sbic.
For \esnli, T5 is better, presumably because \unifiedqa has lost some useful information from \textsc{NLI} after extensive continued pretraining for QA.  
These results suggest that \unifiedqa is a better model for prompting self-rationalization with QA prompts.

To recap, the analysis presented in this section suggests that QA prompting for inducing self-rationalization behavior is best done when \unifiedqa is combined with the NL prompt below. For true QA tasks, we use the original \unifiedqa formats.\footnote{Following \citet{Hendrycks2021MeasuringMM}, we add </s> to the end of our \unifiedqaprompt prompts.}  

\begin{tcolorbox}[width=\columnwidth, colframe=blue!20, colback=blue!10, halign=left, boxsep=0pt, left=6pt,right=6pt]
\textbf{Input:} 

\texttt{explain what is this/more...? \textbackslash\textbackslash n tag$_1$:} [\emph{sequence}$_1$] \texttt{tag$_2$:} [\emph{sequence}$_2$] ...\texttt{</s>}

\vspace{2mm}

\textbf{Output:} 

[\emph{answer/label}] \texttt{because} [\emph{explanation}]
\end{tcolorbox}

\begin{table}[!t]
\centering
\resizebox{\columnwidth}{!}{  
\begin{tabular}{llrr}
\toprule
                  & \textbf{Prompt}                                        & \textbf{Accuracy}  & \textbf{BERTscore}\\
\midrule
\multirow{9}{*}{\rotatebox{90}{\textsc{E-SNLI}}} & \textsc{\small UniFew}                                   & 61.7$_{0.6}$ & 55.8$_{0.5}$  \\
                        & + tags                            & 63.6$_{0.4}$ & 57.3$_{0.4}$  \\
                        \arrayrulecolor{black!20}\cmidrule{2-4}
                        & Is...?  &  47.5$_{0.5}$ & 42.7$_{0.5}$  \\
                        & + tags & 66.6$_{0.5}$ & 60.0$_{0.5}$  \\
                        & + tags \& choices                    & 64.4$_{0.5}$ & 58.2$_{0.5}$  \\
                        \arrayrulecolor{black!20}\cmidrule{2-4}
                        & What is...?              & 40.7$_{0.4}$ & 36.5$_{0.4}$  \\
                        & + tags                   & \textbf{75.0}$_{0.3}$ & \textbf{67.5}$_{0.3}$  \\
                        & + tags \& choices        & 69.3$_{0.7}$ & 62.5$_{0.6}$  \\
                        \arrayrulecolor{black!20}\cmidrule{2-4}
                        & \textsc{\small Random Baseline} & \makecell[c]{33.3} & \makecell[c]{-}\\
\arrayrulecolor{black}\midrule
\multirow{2}{*}{\rotatebox{90}{\ecqa}} & \textsc{\small UnifiedQA}  & \textbf{41.4}$_{0.3}$ & \textbf{36.7}$_{0.3}$ \\
                        \arrayrulecolor{black!20}\cmidrule{2-4}
                        & \textsc{\small Random Baseline} & \makecell[c]{20.0} & \makecell[c]{-}\\
\arrayrulecolor{black}\midrule
\multirow{7}{*}{\rotatebox{90}{ComVE}}  & Is...? & 52.7$_{0.3}$ & 47.7$_{0.3}$ \\
                        & + tags                                         & 52.5$_{0.3}$ & 47.5$_{0.3}$  \\
                        & + tags \& choices                        & 52.2$_{0.3}$ & 47.3$_{0.3}$  \\
                        \arrayrulecolor{black!20}\cmidrule{2-4}
                        & What is...? & 50.6$_{0.2}$ & 45.7$_{0.2}$ \\
                        & + tags                                     & \textbf{67.3}$_{0.7}$ & \textbf{61.0}$_{0.6}$  \\
                        & + tags \& choices                          & 62.6$_{0.6}$ & 56.7$_{0.6}$  \\
                        \arrayrulecolor{black!20}\cmidrule{2-4}
                        & \textsc{\small Random Baseline} & \makecell[c]{50.0} & \makecell[c]{-}\\
\arrayrulecolor{black}\midrule
\multirow{8}{*}{\rotatebox{90}{SBIC}}   & \textsc{\small UniFew} & 66.1$_{0.4}$ & 63.8$_{0.4}$  \\
            \arrayrulecolor{black!20}\cmidrule{2-4}
                        & Is...? & 63.5$_{0.4}$ & 61.2$_{0.4}$  \\
                        & + tags & 62.6$_{0.4}$ & 60.4$_{0.4}$ \\
                        & + tags \& choices &  63.6$_{0.4}$ & 61.3$_{0.4}$ \\
                        \arrayrulecolor{black!20}\cmidrule{2-4}
                        & What is...? & 67.3$_{0.4}$ & 65.0$_{0.4}$  \\
                        & + tags & \textbf{67.5}$_{0.4}$ &	\textbf{65.3}$_{0.4}$ \\
                        & + tags \& choices & 65.4$_{0.6}$ & 63.1$_{0.6}$ \\
                        \arrayrulecolor{black!20}\cmidrule{2-4}
                        & \textsc{\small Random Baseline} & \makecell[c]{50.0} & \makecell[c]{-}\\
\arrayrulecolor{black}\bottomrule
\end{tabular}}
\caption{Prompting \unifiedqa with \unifiedqaprompt with  ``Is...?'' and ``What is...?'' questions, and \textsc{\small UniFew}. See \sect{sec:qa_prompts} for descriptions of these prompts. For \ecqa we use the original \unifiedqa format for multiple-choice QA. We also inspect the effects of adding label choices and \emph{tags} (defined in \sect{sec:qa_prompts}) to the input. 
}
\label{tab:qa_prompts}
\end{table}

\begin{table}[t]
\centering
\begin{subtable}{\columnwidth}
    \resizebox{\columnwidth}{!}{  
        \begin{tabular}{lcccc}
        \toprule      
                  & \esnli          & \ecqa          & \comve          & \sbic           \\
        \midrule
        B    & \textbf{75.2}$_{0.4}$ & 22.3$_{0.3}$  & \textbf{50.4}$_{0.3}$ & 61.6$_{0.4}$ \\
        N & 75.1$_{0.4}$ &  \textbf{27.6}$_{0.4}$ & 49.0$_{0.3}$ & \textbf{64.7}$_{0.5}$ \\
        \bottomrule
        \end{tabular}
    }
    \caption{Accuracy.}
    \end{subtable}\par%
    \vspace*{0.5em}
    \begin{subtable}{\columnwidth}
        \resizebox{\columnwidth}{!}{  
            \begin{tabular}{lcccc}
            \toprule     
                & \esnli          & \ecqa          & \comve          & \sbic           \\
                \midrule
                B   & \textbf{67.7}$_{0.3}$ & 19.8$_{0.3}$ & \textbf{45.5}$_{0.3}$ & 59.2$_{0.5}$ \\
                N & 67.5$_{0.4}$ & \textbf{24.5}$_{0.3}$ & 44.3$_{0.3}$ & \textbf{62.0}$_{0.5}$ \\
                \bottomrule
            \end{tabular}
        }
    \caption{BERTscore.}
    \end{subtable}    
\caption{A comparison of the basic infilling prompt (B; ``\texttt{<extra\_id\_0> because <extra\_id\_1>}'') with its more natural sounding version (N; see \sect{sec:infilling_prompts}). 
}
\label{tab:infilling_prompts}
\end{table}

\subsection{\textsc{Infilling} Prompts }
\label{sec:infilling_prompts}

The simplest way to design an infilling prompt for self-rationalization with T5 is to add the span ``\texttt{<extra\_id\_0> because <extra\_id\_1>}'' to the input. %
A model should then replace \texttt{<extra\_id\_0>} with a label/answer and  \texttt{<extra\_id\_1>} with an explanation. %
Besides being similar to T5's span infilling pretraining task, another benefit of this prompt is that it is very flexible---the span above can be added to any task input. %
This basic infilling prompt could be easily made more natural by prepending phrases such as: ``\texttt{The answer is}'' (\ecqa), ``\texttt{Less common is}'' (\comve), or ``\texttt{This is}'' (\esnli, \sbic). %
We hypothesize that these additional phrases could be beneficial because they suggest  which subset of the vocabulary is the right word for filling in  \texttt{<extra\_id\_0>}. %
We test whether it is beneficial to make the infilling prompt more natural-sounding. % 

\paragraph{Results} T5 results are shown in Table \ref{tab:infilling_prompts}. 
The outcome is mixed---while we observe notable benefits for \ecqa/\sbic, for \esnli/\comve there is a minor difference in favor of the basic prompt. %
A way to explain this is that T5 learned about NLI labels from \textsc{MNLI} during pretraining, so it does not an need additional phrase to nudge it in the right direction. %
\comve results are comparable to the random performance, and the model could not learn the task from the infilling prompt, with or without the additional phrases. 
Thus, we recommend using the more natural version as it is not detrimental to \esnli/\comve performance while it leads to big improvements for \ecqa/\sbic. 

\begin{table}[t]
\centering
 \resizebox{0.825\columnwidth}{!}{  
\begin{tabular}{llrr}
\toprule
             &   \textbf{Task}      & \textbf{Accuracy} & \textbf{BERTscore} \\
\midrule
\multirow{5}{*}{\rotatebox{90}{\infilling}} & \esnli  & 75.1$_{0.4}$ & 67.5$_{0.4}$ \\
                           & \ecqa   &  27.6$_{0.4}$ &  24.5$_{0.3}$ \\
                           & \comve   & 49.0$_{0.3}$ & 44.3$_{0.3}$ \\
                           & \sbic    & 64.7$_{0.5}$ & 62.0$_{0.5}$ \\
                           \arrayrulecolor{black!20}\cmidrule{2-4}
                           & Average & \makecell[c]{54.1} & \makecell[c]{49.6}  \\
\arrayrulecolor{black}\midrule
\multirow{5}{*}{\rotatebox{90}{\simtfive}}       & \esnli  & \textbf{79.2}$_{0.3}$ & \textbf{71.3}$_{0.3}$ \\
                           & \ecqa   & 38.3$_{0.3}$  & 33.9$_{0.3}$   \\
                           & \comve   & 55.9$_{0.3}$ & 50.4$_{0.3}$ \\
                           & \sbic    & 65.1$_{0.6}$ & 62.8$_{0.6}$ \\
                           \arrayrulecolor{black!20}\cmidrule{2-4}
                           & Average & \makecell[c]{59.6} & \makecell[c]{54.6}      \\
\arrayrulecolor{black}\midrule
\multirow{5}{*}{\rotatebox{90}{\unifiedqaprompt}} & \esnli  & 75.0$_{0.3}$ & 67.5$_{0.3}$ \\
                           & \ecqa   & \textbf{41.4}$_{0.3}$  & \textbf{36.7}$_{0.3}$  \\
                           & \comve   & \textbf{67.3}$_{0.7}$ & \textbf{61.0}$_{0.6}$ \\
                           & \sbic    & \textbf{67.5}$_{0.4}$ & \textbf{65.3}$_{0.4}$ \\
                           \arrayrulecolor{black!20}\cmidrule{2-4}
                           & Average & \makecell[c]{\textbf{62.8}} & \makecell[c]{\textbf{57.6}}  \\
\arrayrulecolor{black}\bottomrule
\end{tabular}}
\caption{A comparison between three prompt types: \infilling, \simtfive, and \unifiedqaprompt prompts. See \sect{sec:study_various_prompts} for descriptions of these prompts.}
\label{tab:main_comparison}
\end{table}
\newcolumntype{C}{@{\extracolsep{3cm}}c@{\extracolsep{0pt}}}%

\begin{table*}[!h]
\centering
 \resizebox{\textwidth}{!}{  
\begin{tabular}{ll|rr|rrrrrrrr}
\toprule
& & & &  \multicolumn{8}{c}{\textbf{Plausibility}} \\
& & & &  \multicolumn{2}{c}{\emph{All}} & \multicolumn{2}{c}{\emph{Label}$_1$} & \multicolumn{2}{c}{\emph{Label}$_2$} & \multicolumn{2}{c}{\emph{Label}$_3$} \\    
\cmidrule(l{2pt}r{2pt}){5-6}\cmidrule(l{2pt}r{2pt}){7-8}\cmidrule(l{2pt}r{2pt}){9-10}\cmidrule(l{2pt}r{2pt}){11-12}
& \makecell[c]{\textbf{Model}}  &  \makecell[c]{\textbf{Accuracy}}  & \makecell[c]{\textbf{BERTscore}} & \makecell[c]{\textbf{Score}} & \makecell[c]{\textbf{$\kappa$}} & \makecell[c]{\textbf{Score}} & \makecell[c]{\textbf{$\kappa$}} & \makecell[c]{\textbf{Score}} & \makecell[c]{\textbf{$\kappa$}}  & \makecell[c]{\textbf{Score}} & \makecell[c]{\textbf{$\kappa$}} \\
\midrule
\multirow{6}{*}{\rotatebox{90}{\textsc{E-SNLI}}} & \textsc{\small Base}  & 79.2$_{0.3}$ & 71.3$_{0.3}$ & 16.7$_{1.5}$ & 0.73 &  15.6$_{2.3}$ & 0.67 & 17.5$_{2.9}$ & 0.79 & 17.1$_{2.7}$ & 0.72 \\
                      &\textsc{\small Large}  & 84.8$_{0.3}$ &	76.6$_{0.3}$ & 32.7$_{1.9}$  & 0.57 & \textbf{27.3}$_{2.9}$ & 0.43  & 33.9$_{3.4}$ & 0.64  & 36.8$_{3.6}$ & 0.64  \\
                      &\textsc{\small 3B}  &\textbf{87.4}$_{0.2}$ & \textbf{79.1}$_{0.2}$ & 41.6$_{2.1}$  & 0.62 & 27.1$_{2.8}$& 0.52  & 46.8$_{3.8}$ & 0.70   & \textbf{50.9}$_{3.6}$ & 0.64  \\
\arrayrulecolor{black!20}\cmidrule{2-12}
                      &\textsc{\small GPT-3}  & 65.4$_{0.5}$ & 59.8$_{0.5}$  & \textbf{42.4}$_{2.2}$  & 0.54 & \textbf{27.3}$_{2.9}$ & 0.48 & \textbf{66.0}$_{4.4}$ & 0.71  & 43.8$_{3.5}$ & 0.51 \\
\arrayrulecolor{black!20}\cmidrule{2-12}
&\textsc{\small Gold} &  \cellcolor{black!5} & \cellcolor{black!5} & 77.4$_{1.6}$   & 0.63 & 63.5$_{3.0}$ & 0.44 & 87.9$_{1.8}$ & 0.74  & 82.5$_{2.4}$ & 0.72\\
&\textsc{\small Rand} &  \makecell[c]{33.3} &\cellcolor{black!5}&  \cellcolor{black!5}  & \cellcolor{black!5} & \cellcolor{black!5} &\cellcolor{black!5}  &\cellcolor{black!5}  &\cellcolor{black!5} & \cellcolor{black!5}  & \cellcolor{black!5} \\
\arrayrulecolor{black}\midrule
\multirow{6}{*}{\rotatebox{90}{\textsc{ECQA}}} & \textsc{\small Base}   & 41.4$_{0.3}$ & 36.7$_{0.3}$ & 25.5$_{1.2}$ & 0.32
& \cellcolor{black!5} & \cellcolor{black!5} & \cellcolor{black!5} & \cellcolor{black!5} & \cellcolor{black!5} & \cellcolor{black!5}\\
                      &\textsc{\small Large}  & 57.2$_{0.4}$ &	51.0$_{0.3}$ &  30.3$_{1.5}$ & 0.38  & \cellcolor{black!5} & \cellcolor{black!5} & \cellcolor{black!5} & \cellcolor{black!5} & \cellcolor{black!5} & \cellcolor{black!5}\\
                      &\textsc{\small 3B}  & \textbf{65.9}$_{0.4}$ & \textbf{59.0}$_{0.3}$ & 34.2$_{1.6}$ & 0.35   & \cellcolor{black!5} & \cellcolor{black!5} & \cellcolor{black!5} & \cellcolor{black!5} & \cellcolor{black!5} & \cellcolor{black!5}\\
\arrayrulecolor{black!20}\cmidrule{2-12}
                      &\textsc{\small GPT-3}  & 60.6$_{1.5}$ & 54.4$_{1.3}$ & \textbf{45.1}$_{1.4}$  & 0.12 & \cellcolor{black!5} & \cellcolor{black!5} & \cellcolor{black!5} & \cellcolor{black!5} & \cellcolor{black!5} & \cellcolor{black!5}\\
\arrayrulecolor{black!20}\cmidrule{2-12}                      
                    &\textsc{\small Gold} &   \cellcolor{black!5} & \cellcolor{black!5} & 70.9$_{1.5}$ & 0.45  & \cellcolor{black!5} & \cellcolor{black!5} & \cellcolor{black!5} & \cellcolor{black!5} & \cellcolor{black!5} & \cellcolor{black!5}\\
&\textsc{\small Rand} &  \makecell[c]{20.00} & \cellcolor{black!5} & \cellcolor{black!5} &  \cellcolor{black!5}  & \cellcolor{black!5} & \cellcolor{black!5} & \cellcolor{black!5} & \cellcolor{black!5} & \cellcolor{black!5} & \cellcolor{black!5}\\
\arrayrulecolor{black}\midrule
\multirow{6}{*}{\rotatebox{90}{\textsc{ComVE}}} & \textsc{\small Base}   & 67.3$_{0.7}$ &	61.0$_{0.6}$ & 13.8$_{1.3}$ & 0.45 & \cellcolor{black!5} & \cellcolor{black!5} & \cellcolor{black!5} & \cellcolor{black!5} & \cellcolor{black!5} & \cellcolor{black!5}\\
                     & \textsc{\small Large} & 81.3$_{0.4}$ &	73.9$_{0.4}$ & 25.6$_{1.7}$  & 0.52 & \cellcolor{black!5} & \cellcolor{black!5} & \cellcolor{black!5} & \cellcolor{black!5} & \cellcolor{black!5} & \cellcolor{black!5}\\
                     & \textsc{\small 3B}  & \textbf{89.0}$_{0.4}$ & \textbf{81.0}$_{0.3}$ & 33.4$_{1.7}$ & 0.63 & \cellcolor{black!5} & \cellcolor{black!5} & \cellcolor{black!5} & \cellcolor{black!5} & \cellcolor{black!5} & \cellcolor{black!5}\\
\arrayrulecolor{black!20}\cmidrule{2-12}
                      &\textsc{\small GPT-3}  & 74.0$_{1.4}$ & 67.6$_{1.3}$ & \textbf{42.2}$_{1.8}$ & 0.73 & \cellcolor{black!5} & \cellcolor{black!5} & \cellcolor{black!5} & \cellcolor{black!5} & \cellcolor{black!5} & \cellcolor{black!5}\\
\arrayrulecolor{black!20}\cmidrule{2-12}
                      &\textsc{\small Gold}   & \cellcolor{black!5} & \cellcolor{black!5} & 77.2$_{1.3}$ & 0.55  & \cellcolor{black!5} & \cellcolor{black!5} & \cellcolor{black!5} & \cellcolor{black!5} & \cellcolor{black!5} & \cellcolor{black!5}\\
&\textsc{\small Rand}   & \makecell[c]{50.0} & \cellcolor{black!5} & \cellcolor{black!5} &  \cellcolor{black!5} & \cellcolor{black!5} & \cellcolor{black!5} & \cellcolor{black!5} & \cellcolor{black!5} & \cellcolor{black!5} & \cellcolor{black!5}\\
\arrayrulecolor{black}\midrule                    
\multirow{6}{*}{\rotatebox{90}{\textsc{SBIC}}} & \textsc{\small Base}    & 67.5$_{0.4}$ &	65.3$_{0.4}$ &  58.0$_{2.2}$ & 0.68  & 21.4$_{2.1}$ & 0.54  & 94.6$_{1.1}$ & 0.82 & \cellcolor{black!5} & \cellcolor{black!5}\\
                     & \textsc{\small Large}  & 71.1$_{0.4}$ &	68.5$_{0.4}$ & 61.8$_{2.2}$  & 0.66 & 27.2$_{2.2}$ & 0.43 & \textbf{96.5}$_{0.9}$  & 0.89 & \cellcolor{black!5} & \cellcolor{black!5}\\
                     & \textsc{\small 3B}  & 71.7$_{0.5}$ & 68.9$_{0.5}$ & 64.2$_{2.1}$ & 0.68 & 33.8$_{2.6}$  & 0.55 & 94.6$_{1.0}$ & 0.81 & \cellcolor{black!5} & \cellcolor{black!5}\\
                    \arrayrulecolor{black!20}\cmidrule{2-12}
                    &\textsc{\small GPT-3}  &  \textbf{74.2}$_{1.4}$ & \textbf{71.5}$_{1.4}$ & \textbf{72.7}$_{1.7}$  & 0.53 & \textbf{52.6}$_{2.5}$ & 0.34  & 92.7$_{1.0}$ & 0.72 & \cellcolor{black!5} & \cellcolor{black!5}\\
\arrayrulecolor{black!20}\cmidrule{2-12}
                    &\textsc{\small Gold}   & \cellcolor{black!5} & \cellcolor{black!5} & 79.8$_{1.6}$  & 0.67 & 64.9$_{2.7}$ & 0.52 & 94.7$_{1.0}$ & 0.81 &  \cellcolor{black!5} & \cellcolor{black!5}\\
&\textsc{\small Rand}   & \makecell[c]{50.0} & \cellcolor{black!5} & \cellcolor{black!5} &  \cellcolor{black!5} & \cellcolor{black!5} & \cellcolor{black!5} & \cellcolor{black!5} & \cellcolor{black!5} & \cellcolor{black!5} & \cellcolor{black!5}\\
\arrayrulecolor{black}\bottomrule
\end{tabular}}
\caption{The first results on the \benchmark benchmark using T5/\unifiedqa (\textsc{base}, \textsc{large}, 3B) and \textsc{GPT-3}. T5 with \simtfive prompt is used only for \esnli, and \unifiedqa+\unifiedqaprompt prompt is used for other datasets. The descriptions of these prompts are given in \sect{sec:study_various_prompts} and details of how evaluation metrics are calculated in \sect{sec:benchmark}. \textsc{Rand} stands for a random baseline and \textsc{Gold} for human-authored explanations. \emph{Label}$_1$/\emph{Label}$_2$/\emph{Label}$_3$ are entailment/neutral/contradiction in \esnli and offensive/not offensive in \sbic. The number of parameters is: 200M (\textsc{Base}), 770M (\textsc{Large}), 2.8B (3B), and 175B (GPT-3). 
}
\label{tab:scale_size}
\end{table*}

\subsection{\textsc{Infilling} vs.\ $\approx$T5 vs.\ QA}
\label{sec:main_comparison}

We have established appropriate QA and \textsc{Infilling} prompts in \sect{sec:qa_prompts} and \sect{sec:infilling_prompts}. 
We now turn to a comparison between all three prompt types: (i) \infilling (natural), (ii) \simtfive, and (iii) \unifiedqaprompt (``What is...?'' with tags). 
The first two are used to prompt T5 and the last type \unifiedqa. 
To construct \simtfive prompts, for each task in \benchmark, we identify the most similar T5's pretraining task (see Table \ref{tab:similar_tasks}, Appendix) and use that task's formatting (see, e.g.,
\simtfive $\times$ \textsc{COPA} in Table \ref{tab:prompts_comve}). 

\paragraph{Results} A comparison of the three prompt types is presented in Table \ref{tab:main_comparison}. The \unifiedqaprompt prompt outperforms other prompt types for all tasks except \esnli for which unsurprisingly \simtfive is the best. 
Finally, this brings us to the end of our extensive exploration of natural language prompts for a prompt-based finetuning approach to few-shot self-rationalization. We identify the \unifiedqaprompt prompt as the most effective and we use it to study how few-shot self-rationalization performance scales with the size of the \unifiedqa model.

\section{Improving Self-Rationalization with Increasing Model Size}
\label{sec:scaling}

In \sect{sec:study_various_prompts}, we discovered that a QA prompt combined with the base \unifiedqa model version is as an effective combination for few-shot self-rationalization through prompt-based finetuning. 
In this section, we provide two additional evaluations to establish the first approach to few-shot self-rationalization. 

First, we assess how plausible the generated explanations are when evaluated by annotators on Amazon MTurk. 
Details of how we conduct human evaluation of plausibility are given in \sect{sec:benchmark}. 
One HIT contains 10 instances and we pay \$1 per HIT. 

Next, we investigate how self-rationalization performance changes with the model size since larger pretrained language models typically give better few-shot performance \cite{brown2020gpt3}. 
We wonder whether the same trend will hold for a complex generation task of self-rationalization where it is conceivable that an enormous model could overfit on a few examples. 
To this end, we evaluate three versions of \unifiedqa (\textsc{base}, \textsc{large}, 3B) and \textsc{GPT-3}  \cite{brown2020gpt3}. %
We use \texttt{davinci-instruct-beta} which is a beta version of the \textsc{InstructGPT} model \cite{Ouyang2022TrainingLM}.

We evaluate \textsc{GPT-3} using its API and ``in-context demonstrations''~\cite{brown2020gpt3}. %
We pack as many training examples (demonstrations)  as we can fit in the input, followed by the input of the test example, then run GPT-3 to generate its output. 
The number of demonstrations we are able to fit ranges from [28,45]  which are randomly selected from the 48 used for \unifiedqa. 
Since evaluation using a single prompt costs us \$1,050, we do not do prompt search for \textsc{GPT-3}. 
We use the prompts shown in Fig.~\ref{fig:gpt3_prompts} in Appendix. 

A detailed description of evaluation metrics is given in \sect{sec:benchmark}. The dev set size (of each out of 60 dev sets)  for \textsc{GPT-3} is 18 instead of 350 (because of the API cost). 
Ground-truth explanations are evaluated together with explanations generated by 4 models. %
Therefore, for \textsc{gold} explanations, we report the average of 4 plausibility scores, std.\ errors, and $\kappa$ values calculated with 4 Mturk batches (corresponding to 4 models).

\subsection{Results} 

Results are shown in Table \ref{tab:scale_size}. Note that we use T5 with the \simtfive prompt for \esnli, and \unifiedqa with \unifiedqaprompt (\sect{sec:study_various_prompts}) for other datasets to establish the best possible performance for each dataset. %  
The exact prompts for each task are given in Appendix \ref{sec:finall_prompts}. 
We observe that all metrics---accuracy, BERTscore, and plausibility---monotonically increase with the model size for all datasets. %
That is, larger models learn to predict task labels and generate explanations from a few examples better. 
\unifiedqa-3B has a higher accuracy/BERTscore than \textsc{GPT-3} for all datasets except \textsc{SBIC}, but \textsc{GPT-3} generates explanations that are notably more plausible. 

The following observations suggest that few-shot self-rationalization is a promising research direction. %
The difference in plausibility scores between the \textsc{Base} and 3B model versions ranges from [6.2, 24.8] (on average 14.8). %
In other words, since it is possible to generate more plausible explanations by only increasing the model size, it is conceivable that further progress could be made with more creative approaches. % 
Next, the plausibility score of the best model (\textsc{GPT-3}) ranges from [42.2, 72.7] ([42.2, 52.6] if we consider only \sbic ``offensive'' (\emph{Label}$_1$) subset. %
This shows that a moderate plausibility can already be achieved with current models without any task-specific enhancements. %  

Despite that, the gap between our best models and human-authored explanations remains  large. %
The average plausibility score across datasets is 43.4 (\unifiedqa-3B), 50.6 (\textsc{GPT-3}), and 76.3 (\textsc{Gold}). %
In other words, the difference in plausibility scores between \unifiedqa-3B's and human explanations is 33.0, and between \textsc{GPT-3}'s and human explanations is 25.7. %
We expect that the \benchmark benchmark, our \unifiedqa approach, and first results, present a good starting point to tackle this challenge.  

\paragraph{Performance w.r.t.\ Labels} For \esnli and \sbic, we can inspect the metrics with respect to labels. %
In \esnli part of the Table \ref{tab:scale_size}, \emph{Label}$_1$ marks ``entailment'', \emph{Label}$_2$  ``neutral'', and \emph{Label}$_3$  ``contradiction''. %
There are notable differences between the plausibility scores for each label. %
The plausibility score for ``entailment'' does not scale with the model size and it is much lower than scores for other labels (the best score is  27.3 vs.\ 66.0/50.9). %
This issue stems from the difficulty of explaining the entailment label \cite{camburu2018snli}. 
Even people struggle with explaining ``entailment'' as evident by the lower \textsc{gold} score for ``entailment'' compared to the other two labels. %
An interesting observation from the other two labels is that \unifiedqa-3B is explains  ``contradiction'' instances best and \textsc{GPT-3} ``neutral'' instances. %

In \sbic part of the Table \ref{tab:scale_size}, \emph{Label}$_1$ marks ``offensive'' and \emph{Label}$_2$ ``not offensive'' instances. %
The latter achieve almost perfect plausibility since the models learn to generated ``\emph{This post does not imply anything offensive}''. Thus, main plausibility scores for \sbic are those of offensive instances. %
We can observe that the relative differences between models for offensive instances are much larger than the relative differences when examples of both labels are accounted for (column ``\emph{All} / Score''). If we had only looked into a single plausibility score we would not notice these differences. %
This result is in line with \citet{carton-etal-2020-evaluating} who also recommend breaking down the evaluation of explanations w.r.t. labels whenever possible. % 

\paragraph{Annotator Agreement} Finally, we observe challenges in collecting human judgments of plausibility. 
For all datasets except \ecqa, Fleiss' $\kappa$ is either moderate (between 0.41--0.6) or substantial (between 0.61--0.8). 
One exception is \textsc{GPT-3} on \sbic (\emph{Label}$_1$; offensive) where $\kappa$ is only 0.34. 
We also observe  that $\kappa$ for \textsc{GPT-3}'s explanations is lower than  
$\kappa$ for \unifiedqa's or \textsc{gold} explanations, with the exception of \comve. 
The most concerning is \ecqa where $\kappa$ is on average 0.35 for \unifiedqa's explanations, 0.34 for \textsc{Gold} explanations, and only 0.12 for \textsc{GPT-3}'s. 
Future work should investigate the reasons behind these differences more carefully.  

\section{Related Work}

\paragraph{Few-Shot Self-Rationalization} A standard approach to creating explanations in the form of \emph{highlights} is the select-then-predict method \cite{lei-etal-2016-rationalizing} that  does not use any human-author input highlights. %
On the other hand, a standard method for generating free-text explanations is to use human-written explanations \cite[among others]{liu-etal-2019-towards-explainable, wu-mooney-2019-faithful, narang2020wt5}. %
To the best of our knowledge, prior to submitting our work only two prior works have generated free-text explanations in a weakly-supervised way from the task prediction loss. 
\citet{latcinnik2020explaining} approach commonsense QA in that fashion. %
\citet{Brahman2021LearningTR} propose a distant supervision approach to explaining a defeasible inference task. %
In this paper, we introduce the \benchmark benchmark to unify the evaluation of few-shot self-rationalization and present the first approach and results on \benchmark. 

Concurrent to our work, \citet{Yordanov2021FewShotOT} study  self-rationalization transfer from a high-resource task to a task with only a few human-authored explanations. %
\citet{wiegreffe2021reframing} analyze explanations obtained by  prompting GPT-3 multiple times to get multiple explanation candidates, and then filter these candidates using a model trained to predict acceptability of explanations. Their prompt consists of a few examples with high-quality explanations written by the authors and a new instance together with its gold label. %
\citet{Wei2022ChainOT} demonstrate end-task performance improvements  attained by prompting the PaLM model \cite{Chowdhery2022PaLMSL} to first generate an explanation behind its reasoning (``chain of thought'') and then the task label. % 
\citet{Zelikman2022STaRBR} extend this approach by using  explanations generated in a few-shot manner to refine the same  GPT-J \cite{gpt-j} model. 

\paragraph{Few-Shot Learning} We study natural language prompts \cite{brown2020gpt3, schick-schutze-2021-exploiting} to establish the first approach to few-shot self-rationalization. %
Alternatively, few-shot learning researchers are studying prompts in the form of continuous/soft vectors that do not correspond to real tokens \cite[e.g.,][]{qin-eisner-2021-learning}. Such methods present a promising research direction for few-shot self-rationalization. % 
Namely, we show that larger models generate notably more plausible explanations, and ``prefix tuning'' \cite{li-liang-2021-prefix} has been show to learn two condition generation tasks using only 0.1\% of the parameters, while maintaining comparable performance. % 
In practice, such approaches still require a notable amount of GPU memory. %
Thus, any efforts to reduce required memory such as compression \cite{ganesh_tacl_a_00413} may be valuable for few-shot self-rationalization. 

\section{Conclusions}

We draw attention to the task of few-shot self-rationalization: predicting task labels and generating \emph{free-text} explanations for the prediction using only a few human-written explanations. % 
We present (i) the \benchmark benchmark, (ii) the first prompting approach for \benchmark established through a comprehensive search of natural language prompts, and (iii) results using models with a number of parameters ranging from 220M to 175B. %
Our human evaluation results show that progress is possible on this task given that just scaling the model size increases both the  plausibility of generated explanations and task accuracy by a very large margin. %
Despite that, few-shot self-rationalization remains very challenging, with the plausibility of explanations generated by the best model being 27.7  points behind that of human-authored explanations. %
We hope that work presented in this paper spurs the community to work on this challenging problem to enable more intuitive interaction with NLP systems. 

\section*{Acknowledgments}
The authors thank members of the AllenNLP team and anonymous reviewers for helpful feedback.

\bibliography{anthology,custom}

\begin{thebibliography}{47}
\expandafter\ifx\csname natexlab\endcsname\relax\def\natexlab#1{#1}\fi

\bibitem[{Aggarwal et~al.(2021)Aggarwal, Mandowara, Agrawal, Khandelwal,
  Singla, and Garg}]{aggarwal-etal-2021-explanations}
Shourya Aggarwal, Divyanshu Mandowara, Vishwajeet Agrawal, Dinesh Khandelwal,
  Parag Singla, and Dinesh Garg. 2021.
\newblock \href {https://doi.org/10.18653/v1/2021.acl-long.238} {{E}xplanations
  for {C}ommonsense{QA}: {N}ew {D}ataset and {M}odels}.
\newblock In \emph{Proceedings of the 59th Annual Meeting of the Association
  for Computational Linguistics and the 11th International Joint Conference on
  Natural Language Processing (Volume 1: Long Papers)}, pages 3050--3065,
  Online. Association for Computational Linguistics.

\bibitem[{Bragg et~al.(2021)Bragg, Cohan, Lo, and Beltagy}]{Bragg2021FLEXUE}
Jonathan Bragg, Arman Cohan, Kyle Lo, and Iz~Beltagy. 2021.
\newblock \href {https://arxiv.org/abs/2107.07170} {Flex: Unifying evaluation
  for few-shot nlp}.
\newblock In \emph{Proceedings of the Advances in Neural Information Processing
  Systems (NeurIPS)}.

\bibitem[{Brahman et~al.(2021)Brahman, Shwartz, Rudinger, and
  Choi}]{Brahman2021LearningTR}
Faeze Brahman, Vered Shwartz, Rachel Rudinger, and Yejin Choi. 2021.
\newblock \href {https://arxiv.org/abs/2012.08012} {Learning to rationalize for
  nonmonotonic reasoning with distant supervision}.
\newblock In \emph{Proceedings of the AAAI Conference on Artificial
  Intelligence}.

\bibitem[{Brown et~al.(2020)Brown, Mann, Ryder, Subbiah, Kaplan, Dhariwal,
  Neelakantan, Shyam, Sastry, Askell, Agarwal, Herbert-Voss, Krueger, Henighan,
  Child, Ramesh, Ziegler, Wu, Winter, Hesse, Chen, Sigler, Litwin, Gray, Chess,
  Clark, Berner, McCandlish, Radford, Sutskever, and Amodei}]{brown2020gpt3}
Tom Brown, Benjamin Mann, Nick Ryder, Melanie Subbiah, Jared~D Kaplan, Prafulla
  Dhariwal, Arvind Neelakantan, Pranav Shyam, Girish Sastry, Amanda Askell,
  Sandhini Agarwal, Ariel Herbert-Voss, Gretchen Krueger, Tom Henighan, Rewon
  Child, Aditya Ramesh, Daniel Ziegler, Jeffrey Wu, Clemens Winter, Chris
  Hesse, Mark Chen, Eric Sigler, Mateusz Litwin, Scott Gray, Benjamin Chess,
  Jack Clark, Christopher Berner, Sam McCandlish, Alec Radford, Ilya Sutskever,
  and Dario Amodei. 2020.
\newblock \href
  {https://proceedings.neurips.cc/paper/2020/file/1457c0d6bfcb4967418bfb8ac142f64a-Paper.pdf}
  {Language models are few-shot learners}.
\newblock In \emph{Advances in Neural Information Processing Systems},
  volume~33, pages 1877--1901. Curran Associates, Inc.

\bibitem[{Camburu et~al.(2018)Camburu, Rockt{\"a}schel, Lukasiewicz, and
  Blunsom}]{camburu2018snli}
Oana-Maria Camburu, Tim Rockt{\"a}schel, Thomas Lukasiewicz, and Phil Blunsom.
  2018.
\newblock \href
  {https://papers.nips.cc/paper/8163-e-snli-natural-language-inference-with-natural-language-explanations/}
  {{e-SNLI: N}atural language inference with natural language explanations}.
\newblock In \emph{Proceedings of the Advances in Neural Information Processing
  Systems (NeurIPS)}.

\bibitem[{Carton et~al.(2020)Carton, Rathore, and
  Tan}]{carton-etal-2020-evaluating}
Samuel Carton, Anirudh Rathore, and Chenhao Tan. 2020.
\newblock \href {https://doi.org/10.18653/v1/2020.emnlp-main.747} {Evaluating
  and characterizing human rationales}.
\newblock In \emph{Proceedings of the 2020 Conference on Empirical Methods in
  Natural Language Processing (EMNLP)}, pages 9294--9307, Online. Association
  for Computational Linguistics.

\bibitem[{Chowdhery et~al.(2022)Chowdhery, Narang, Devlin, Bosma, Mishra,
  Roberts, Barham, Chung, Sutton, Gehrmann, Schuh, Shi, Tsvyashchenko, Maynez,
  Rao, Barnes, Tay, Shazeer, Prabhakaran, Reif, Du, Hutchinson, Pope, Bradbury,
  Austin, Isard, Gur-Ari, Yin, Duke, Levskaya, Ghemawat, Dev, Michalewski,
  Garc{\'i}a, Misra, Robinson, Fedus, Zhou, Ippolito, Luan, Lim, Zoph,
  Spiridonov, Sepassi, Dohan, Agrawal, Omernick, Dai, Pillai, Pellat,
  Lewkowycz, Moreira, Child, Polozov, Lee, Zhou, Wang, Saeta, Diaz, Firat,
  Catasta, Wei, Meier-Hellstern, Eck, Dean, Petrov, and
  Fiedel}]{Chowdhery2022PaLMSL}
Aakanksha Chowdhery, Sharan Narang, Jacob Devlin, Maarten Bosma, Gaurav Mishra,
  Adam Roberts, Paul Barham, Hyung~Won Chung, Charles Sutton, Sebastian
  Gehrmann, Parker Schuh, Kensen Shi, Sasha Tsvyashchenko, Joshua Maynez,
  Abhishek~Baindoor Rao, Parker Barnes, Yi~Tay, Noam~M. Shazeer, Vinodkumar
  Prabhakaran, Emily Reif, Nan Du, Benton~C. Hutchinson, Reiner Pope, James
  Bradbury, Jacob Austin, Michael Isard, Guy Gur-Ari, Pengcheng Yin, Toju Duke,
  Anselm Levskaya, Sanjay Ghemawat, Sunipa Dev, Henryk Michalewski, Xavier
  Garc{\'i}a, Vedant Misra, Kevin Robinson, Liam Fedus, Denny Zhou, Daphne
  Ippolito, David Luan, Hyeontaek Lim, Barret Zoph, Alexander Spiridonov, Ryan
  Sepassi, David Dohan, Shivani Agrawal, Mark Omernick, Andrew~M. Dai,
  Thanumalayan~Sankaranarayana Pillai, Marie Pellat, Aitor Lewkowycz,
  Erica~Oliveira Moreira, Rewon Child, Oleksandr Polozov, Katherine Lee,
  Zongwei Zhou, Xuezhi Wang, Brennan Saeta, Mark Diaz, Orhan Firat, Michele
  Catasta, Jason Wei, Kathleen~S. Meier-Hellstern, Douglas Eck, Jeff Dean, Slav
  Petrov, and Noah Fiedel. 2022.
\newblock \href {https://arxiv.org/abs/2204.02311} {Palm: Scaling language
  modeling with pathways}.
\newblock {arXiv:2204.02311}.

\bibitem[{Devlin et~al.(2019)Devlin, Chang, Lee, and
  Toutanova}]{devlin-etal-2019-bert}
Jacob Devlin, Ming-Wei Chang, Kenton Lee, and Kristina Toutanova. 2019.
\newblock \href {https://doi.org/10.18653/v1/N19-1423} {{BERT}: Pre-training of
  deep bidirectional transformers for language understanding}.
\newblock In \emph{Proceedings of the 2019 Conference of the North {A}merican
  Chapter of the Association for Computational Linguistics: Human Language
  Technologies, Volume 1 (Long and Short Papers)}, pages 4171--4186,
  Minneapolis, Minnesota. Association for Computational Linguistics.

\bibitem[{Ganesh et~al.(2021)Ganesh, Chen, Lou, Khan, Yang, Sajjad, Nakov,
  Chen, and Winslett}]{ganesh_tacl_a_00413}
Prakhar Ganesh, Yao Chen, Xin Lou, Mohammad~Ali Khan, Yin Yang, Hassan Sajjad,
  Preslav Nakov, Deming Chen, and Marianne Winslett. 2021.
\newblock \href {https://doi.org/10.1162/tacl_a_00413} {{Compressing
  Large-Scale Transformer-Based Models: A Case Study on BERT}}.
\newblock \emph{Transactions of the Association for Computational Linguistics},
  9:1061--1080.

\bibitem[{Gao et~al.(2021)Gao, Fisch, and Chen}]{gao-etal-2021-making}
Tianyu Gao, Adam Fisch, and Danqi Chen. 2021.
\newblock \href {https://doi.org/10.18653/v1/2021.acl-long.295} {Making
  pre-trained language models better few-shot learners}.
\newblock In \emph{Proceedings of the 59th Annual Meeting of the Association
  for Computational Linguistics and the 11th International Joint Conference on
  Natural Language Processing (Volume 1: Long Papers)}, pages 3816--3830,
  Online. Association for Computational Linguistics.

\bibitem[{Gardner et~al.(2019)Gardner, Berant, Hajishirzi, Talmor, and
  Min}]{Gardner2019QuestionAI}
Matt Gardner, Jonathan Berant, Hannaneh Hajishirzi, Alon Talmor, and Sewon Min.
  2019.
\newblock \href {https://arxiv.org/abs/1909.11291} {Question answering is a
  format; when is it useful?}
\newblock {arXiv:1909.11291}.

\bibitem[{Hendrycks et~al.(2021)Hendrycks, Burns, Basart, Zou, Mazeika, Song,
  and Steinhardt}]{Hendrycks2021MeasuringMM}
Dan Hendrycks, Collin Burns, Steven Basart, Andy Zou, Mantas Mazeika, Dawn
  Song, and Jacob Steinhardt. 2021.
\newblock \href {https://arxiv.org/abs/2009.03300} {Measuring massive multitask
  language understanding}.
\newblock In \emph{The International Conference on Learning Representations
  (ICLR)}.

\bibitem[{Jiang et~al.(2020)Jiang, Xu, Araki, and
  Neubig}]{jiang-etal-2020-know}
Zhengbao Jiang, Frank~F. Xu, Jun Araki, and Graham Neubig. 2020.
\newblock \href {https://doi.org/10.1162/tacl_a_00324} {How can we know what
  language models know?}
\newblock \emph{Transactions of the Association for Computational Linguistics},
  8:423--438.

\bibitem[{Kaplan et~al.(2020)Kaplan, McCandlish, Henighan, Brown, Chess, Child,
  Gray, Radford, Wu, and Amodei}]{Kaplan2020ScalingLF}
Jared Kaplan, Sam McCandlish, Tom Henighan, Tom~B. Brown, Benjamin Chess, Rewon
  Child, Scott Gray, Alec Radford, Jeff Wu, and Dario Amodei. 2020.
\newblock \href {https://arxiv.org/abs/2001.08361} {Scaling laws for neural
  language models}.
\newblock {arXiv:2001.08361}.

\bibitem[{Kayser et~al.(2021)Kayser, Camburu, Salewski, Emde, Do, Akata, and
  Lukasiewicz}]{Kayser2021eViLAD}
Maxime Kayser, Oana-Maria Camburu, Leonard Salewski, Cornelius Emde, Virginie
  Do, Zeynep Akata, and Thomas Lukasiewicz. 2021.
\newblock \href
  {https://openaccess.thecvf.com/content/ICCV2021/papers/Kayser_E-ViL_A_Dataset_and_Benchmark_for_Natural_Language_Explanations_in_ICCV_2021_paper.pdf}
  {e-vil: A dataset and benchmark for natural language explanations in
  vision-language tasks}.
\newblock In \emph{Proceedings of the IEEE/CVF International Conference on
  Computer Vision (ICCV)}.

\bibitem[{Khashabi et~al.(2020)Khashabi, Min, Khot, Sabharwal, Tafjord, Clark,
  and Hajishirzi}]{khashabi-etal-2020-unifiedqa}
Daniel Khashabi, Sewon Min, Tushar Khot, Ashish Sabharwal, Oyvind Tafjord,
  Peter Clark, and Hannaneh Hajishirzi. 2020.
\newblock \href {https://doi.org/10.18653/v1/2020.findings-emnlp.171}
  {{UNIFIEDQA}: Crossing format boundaries with a single {QA} system}.
\newblock In \emph{Findings of the Association for Computational Linguistics:
  EMNLP 2020}, pages 1896--1907, Online. Association for Computational
  Linguistics.

\bibitem[{Kim et~al.(2021)Kim, Pavlick, Ayan, and
  Ramachandran}]{Kim2021WhichLI}
Najoung Kim, Ellie Pavlick, Burcu~Karagol Ayan, and Deepak Ramachandran. 2021.
\newblock \href {https://arxiv.org/abs/2101.00391} {Which linguist invented the
  lightbulb? presupposition verification for question-answering}.
\newblock {arXiv:2101.00391}.

\bibitem[{Latcinnik and Berant(2020)}]{latcinnik2020explaining}
Veronica Latcinnik and Jonathan Berant. 2020.
\newblock \href {https://arxiv.org/pdf/2004.05569.pdf} {Explaining question
  answering models through text generation}.
\newblock {arXiv:2004.05569}.

\bibitem[{Lei et~al.(2016)Lei, Barzilay, and
  Jaakkola}]{lei-etal-2016-rationalizing}
Tao Lei, Regina Barzilay, and Tommi Jaakkola. 2016.
\newblock \href {https://doi.org/10.18653/v1/D16-1011} {Rationalizing neural
  predictions}.
\newblock In \emph{Proceedings of the 2016 Conference on Empirical Methods in
  Natural Language Processing}, pages 107--117, Austin, Texas. Association for
  Computational Linguistics.

\bibitem[{Li and Liang(2021)}]{li-liang-2021-prefix}
Xiang~Lisa Li and Percy Liang. 2021.
\newblock \href {https://doi.org/10.18653/v1/2021.acl-long.353} {Prefix-tuning:
  Optimizing continuous prompts for generation}.
\newblock In \emph{Proceedings of the 59th Annual Meeting of the Association
  for Computational Linguistics and the 11th International Joint Conference on
  Natural Language Processing (Volume 1: Long Papers)}, pages 4582--4597,
  Online. Association for Computational Linguistics.

\bibitem[{Liu et~al.(2019)Liu, Yin, and
  Wang}]{liu-etal-2019-towards-explainable}
Hui Liu, Qingyu Yin, and William~Yang Wang. 2019.
\newblock \href {https://doi.org/10.18653/v1/P19-1560} {Towards explainable
  {NLP}: A generative explanation framework for text classification}.
\newblock In \emph{Proceedings of the 57th Annual Meeting of the Association
  for Computational Linguistics}, pages 5570--5581, Florence, Italy.
  Association for Computational Linguistics.

\bibitem[{Marasovi{\'c} et~al.(2020)Marasovi{\'c}, Bhagavatula, Park, Le~Bras,
  Smith, and Choi}]{marasovic-etal-2020-natural}
Ana Marasovi{\'c}, Chandra Bhagavatula, Jae~sung Park, Ronan Le~Bras, Noah~A.
  Smith, and Yejin Choi. 2020.
\newblock \href {https://doi.org/10.18653/v1/2020.findings-emnlp.253} {Natural
  language rationales with full-stack visual reasoning: From pixels to semantic
  frames to commonsense graphs}.
\newblock In \emph{Findings of the Association for Computational Linguistics:
  EMNLP 2020}, pages 2810--2829, Online. Association for Computational
  Linguistics.

\bibitem[{Melis and Jaakkola(2018)}]{melis2018towards}
David~Alvarez Melis and Tommi Jaakkola. 2018.
\newblock \href
  {http://papers.nips.cc/paper/8003-towards-robust-interpretability-with-self-explaining-neural-networks/}
  {Towards robust interpretability with self-explaining neural networks}.
\newblock In \emph{Proceedings of the Advances in Neural Information Processing
  Systems (NeurIPS)}.

\bibitem[{Narang et~al.(2020)Narang, Raffel, Lee, Roberts, Fiedel, and
  Malkan}]{narang2020wt5}
Sharan Narang, Colin Raffel, Katherine Lee, Adam Roberts, Noah Fiedel, and
  Karishma Malkan. 2020.
\newblock \href {https://arxiv.org/abs/2004.14546} {{WT5?! Training
  Text-to-Text Models to Explain their Predictions}}.
\newblock {arXiv:2004.14546}.

\bibitem[{Ouyang et~al.(2022)Ouyang, Wu, Jiang, Almeida, Wainwright, Mishkin,
  Zhang, Agarwal, Slama, Ray, Schulman, Hilton, Kelton, Miller, Simens, Askell,
  Welinder, Christiano, Leike, and Lowe}]{Ouyang2022TrainingLM}
Long Ouyang, Jeff Wu, Xu~Jiang, Diogo Almeida, Carroll~L. Wainwright, Pamela
  Mishkin, Chong Zhang, Sandhini Agarwal, Katarina Slama, Alex Ray, John
  Schulman, Jacob Hilton, Fraser Kelton, Luke~E. Miller, Maddie Simens, Amanda
  Askell, Peter Welinder, Paul~Francis Christiano, Jan Leike, and Ryan~J. Lowe.
  2022.
\newblock \href {https://arxiv.org/abs/2203.02155} {Training language models to
  follow instructions with human feedback}.
\newblock {arXiv:2203.02155}.

\bibitem[{Qin and Eisner(2021)}]{qin-eisner-2021-learning}
Guanghui Qin and Jason Eisner. 2021.
\newblock \href {https://doi.org/10.18653/v1/2021.naacl-main.410} {Learning how
  to ask: Querying {LM}s with mixtures of soft prompts}.
\newblock In \emph{Proceedings of the 2021 Conference of the North American
  Chapter of the Association for Computational Linguistics: Human Language
  Technologies}, pages 5203--5212, Online. Association for Computational
  Linguistics.

\bibitem[{Raffel et~al.(2020)Raffel, Shazeer, Roberts, Lee, Narang, Matena,
  Zhou, Li, and Liu}]{raffel2019exploring}
Colin Raffel, Noam Shazeer, Adam Roberts, Katherine Lee, Sharan Narang, Michael
  Matena, Yanqi Zhou, Wei Li, and Peter~J. Liu. 2020.
\newblock \href {http://jmlr.org/papers/v21/20-074.html} {Exploring the limits
  of transfer learning with a unified text-to-text transformer}.
\newblock \emph{Journal of Machine Learning Research}, 21(140):1--67.

\bibitem[{Rajani et~al.(2019)Rajani, McCann, Xiong, and
  Socher}]{rajani-etal-2019-explain}
Nazneen~Fatema Rajani, Bryan McCann, Caiming Xiong, and Richard Socher. 2019.
\newblock \href {https://doi.org/10.18653/v1/P19-1487} {Explain yourself!
  leveraging language models for commonsense reasoning}.
\newblock In \emph{Proceedings of the 57th Annual Meeting of the Association
  for Computational Linguistics}, pages 4932--4942, Florence, Italy.
  Association for Computational Linguistics.

\bibitem[{Rajpurkar et~al.(2016)Rajpurkar, Zhang, Lopyrev, and
  Liang}]{rajpurkar-etal-2016-squad}
Pranav Rajpurkar, Jian Zhang, Konstantin Lopyrev, and Percy Liang. 2016.
\newblock \href {https://doi.org/10.18653/v1/D16-1264} {{SQ}u{AD}: 100,000+
  questions for machine comprehension of text}.
\newblock In \emph{Proceedings of the 2016 Conference on Empirical Methods in
  Natural Language Processing}, pages 2383--2392, Austin, Texas. Association
  for Computational Linguistics.

\bibitem[{Roemmele et~al.(2011)Roemmele, Bejan, and Gordon}]{Roemmele2011COPA}
Melissa Roemmele, Cosmin~Adrian Bejan, and Andrew~S. Gordon. 2011.
\newblock \href {http://commonsensereasoning.org/2011/papers/Roemmele.pdf}
  {Choice of plausible alternatives: An evaluation of commonsense causal
  reasoning}.
\newblock In \emph{AAAI Spring Symposium Series}.

\bibitem[{Rudin et~al.(2021)Rudin, Chen, Chen, Huang, Semenova, and
  Zhong}]{Rudin2021InterpretableML}
Cynthia Rudin, Chaofan Chen, Zhi Chen, Haiyang Huang, Lesia Semenova, and Chudi
  Zhong. 2021.
\newblock \href {https://arxiv.org/abs/2103.11251} {Interpretable machine
  learning: Fundamental principles and 10 grand challenges}.
\newblock {arXiv: 2103.11251}.

\bibitem[{Sap et~al.(2020)Sap, Gabriel, Qin, Jurafsky, Smith, and
  Choi}]{sap-etal-2020-social}
Maarten Sap, Saadia Gabriel, Lianhui Qin, Dan Jurafsky, Noah~A. Smith, and
  Yejin Choi. 2020.
\newblock \href {https://doi.org/10.18653/v1/2020.acl-main.486} {Social bias
  frames: Reasoning about social and power implications of language}.
\newblock In \emph{Proceedings of the 58th Annual Meeting of the Association
  for Computational Linguistics}, pages 5477--5490, Online. Association for
  Computational Linguistics.

\bibitem[{Schick and Sch{\"u}tze(2021)}]{schick-schutze-2021-exploiting}
Timo Schick and Hinrich Sch{\"u}tze. 2021.
\newblock \href {https://doi.org/10.18653/v1/2021.eacl-main.20} {Exploiting
  cloze-questions for few-shot text classification and natural language
  inference}.
\newblock In \emph{Proceedings of the 16th Conference of the European Chapter
  of the Association for Computational Linguistics: Main Volume}, pages
  255--269, Online. Association for Computational Linguistics.

\bibitem[{Wang and Komatsuzaki(2021)}]{gpt-j}
Ben Wang and Aran Komatsuzaki. 2021.
\newblock {GPT-J-6B: A 6 Billion Parameter Autoregressive Language Model}.
\newblock \url{https://github.com/kingoflolz/mesh-transformer-jax}.

\bibitem[{Wang et~al.(2019)Wang, Liang, Zhang, Li, and
  Gao}]{wang-etal-2019-make}
Cunxiang Wang, Shuailong Liang, Yue Zhang, Xiaonan Li, and Tian Gao. 2019.
\newblock \href {https://doi.org/10.18653/v1/P19-1393} {Does it make sense? and
  why? a pilot study for sense making and explanation}.
\newblock In \emph{Proceedings of the 57th Annual Meeting of the Association
  for Computational Linguistics}, pages 4020--4026, Florence, Italy.
  Association for Computational Linguistics.

\bibitem[{Warstadt et~al.(2019)Warstadt, Singh, and
  Bowman}]{warstadt-etal-2019-neural}
Alex Warstadt, Amanpreet Singh, and Samuel~R. Bowman. 2019.
\newblock \href {https://doi.org/10.1162/tacl_a_00290} {Neural network
  acceptability judgments}.
\newblock \emph{Transactions of the Association for Computational Linguistics},
  7:625--641.

\bibitem[{Wei et~al.(2022)Wei, Wang, Schuurmans, Bosma, Chi, Le, and
  Zhou}]{Wei2022ChainOT}
Jason Wei, Xuezhi Wang, Dale Schuurmans, Maarten Bosma, Ed~Chi, Quoc Le, and
  Denny Zhou. 2022.
\newblock \href {https://arxiv.org/abs/2201.11903} {Chain of thought prompting
  elicits reasoning in large language models}.
\newblock {arXiv:2201.11903}.

\bibitem[{Wiegreffe et~al.(2022)Wiegreffe, Hessel, Swayamdipta, Riedl, and
  Choi}]{wiegreffe2021reframing}
Sarah Wiegreffe, Jack Hessel, Swabha Swayamdipta, Mark Riedl, and Yejin Choi.
  2022.
\newblock \href {https://arxiv.org/abs/2112.08674} {Reframing human-ai
  collaboration for generating free-text explanations}.
\newblock In \emph{Proceedings of the 2022 Conference of the North American
  Chapter of the Association for Computational Linguistics}.

\bibitem[{Wiegreffe and Marasovi\'{c}(2021)}]{wiegreffe-marasovic-2021-review}
Sarah Wiegreffe and Ana Marasovi\'{c}. 2021.
\newblock \href {https://arxiv.org/abs/2102.12060} {Teach me to explain: A
  review of datasets for explainable nlp}.
\newblock In \emph{Proceedings of the Advances in Neural Information Processing
  Systems (NeurIPS)}.

\bibitem[{Wiegreffe et~al.(2021)Wiegreffe, Marasovi{\'c}, and
  Smith}]{wiegreffe-etal-2021-measuring}
Sarah Wiegreffe, Ana Marasovi{\'c}, and Noah~A. Smith. 2021.
\newblock \href {https://aclanthology.org/2021.emnlp-main.804} {{M}easuring
  association between labels and free-text rationales}.
\newblock In \emph{Proceedings of the 2021 Conference on Empirical Methods in
  Natural Language Processing}, pages 10266--10284, Online and Punta Cana,
  Dominican Republic. Association for Computational Linguistics.

\bibitem[{Williams et~al.(2018)Williams, Nangia, and
  Bowman}]{williams-etal-2018-broad}
Adina Williams, Nikita Nangia, and Samuel Bowman. 2018.
\newblock \href {https://doi.org/10.18653/v1/N18-1101} {A broad-coverage
  challenge corpus for sentence understanding through inference}.
\newblock In \emph{Proceedings of the 2018 Conference of the North {A}merican
  Chapter of the Association for Computational Linguistics: Human Language
  Technologies, Volume 1 (Long Papers)}, pages 1112--1122, New Orleans,
  Louisiana. Association for Computational Linguistics.

\bibitem[{Wu and Mooney(2019)}]{wu-mooney-2019-faithful}
Jialin Wu and Raymond Mooney. 2019.
\newblock \href {https://doi.org/10.18653/v1/W19-4812} {Faithful multimodal
  explanation for visual question answering}.
\newblock In \emph{Proceedings of the 2019 ACL Workshop BlackboxNLP: Analyzing
  and Interpreting Neural Networks for NLP}, pages 103--112, Florence, Italy.
  Association for Computational Linguistics.

\bibitem[{Yang et~al.(2019)Yang, Du, and Hu}]{Yang2019EvaluatingEW}
Fan Yang, Mengnan Du, and Xia Hu. 2019.
\newblock \href {https://arxiv.org/abs/1907.06831} {Evaluating explanation
  without ground truth in interpretable machine learning}.
\newblock {1907.06831}.

\bibitem[{Yordanov et~al.(2021)Yordanov, Kocijan, Lukasiewicz, and
  Camburu}]{Yordanov2021FewShotOT}
Yordan Yordanov, Vid Kocijan, Thomas Lukasiewicz, and Oana-Maria Camburu. 2021.
\newblock \href {https://arxiv.org/abs/2112.06204} {Few-shot out-of-domain
  transfer learning of natural language explanations}.
\newblock In \emph{Proceedings of the Deep Generative Models and Downstream
  Applications Workshop at NeurIPS 2021}.

\bibitem[{Zelikman et~al.(2022)Zelikman, Wu, and Goodman}]{Zelikman2022STaRBR}
Eric Zelikman, Yuhuai Wu, and Noah~D. Goodman. 2022.
\newblock \href {https://arxiv.org/abs/2203.14465} {Star: Bootstrapping
  reasoning with reasoning}.
\newblock {arXiv:2203.14465}.

\bibitem[{Zhang et~al.(2018)Zhang, Liu, Liu, Gao, Duh, and
  Durme}]{Zhang2018ReCoRDBT}
Sheng Zhang, Xiaodong Liu, Jingjing Liu, Jianfeng Gao, Kevin Duh, and
  Benjamin~Van Durme. 2018.
\newblock \href {https://arxiv.org/abs/1810.12885} {Record: Bridging the gap
  between human and machine commonsense reading comprehension}.
\newblock {arXiv:1810.12885}.

\bibitem[{Zhang et~al.(2020)Zhang, Kishore, Wu, Weinberger, and
  Artzi}]{Zhang2020BERTScoreET}
Tianyi Zhang, Varsha Kishore, Felix Wu, Kilian~Q. Weinberger, and Yoav Artzi.
  2020.
\newblock \href {https://arxiv.org/abs/1904.09675} {Bertscore: Evaluating text
  generation with bert}.
\newblock In \emph{The International Conference on Learning Representations
  (ICLR)}.

\end{thebibliography}
\bibliographystyle{acl_natbib}

\appendix

\clearpage

\section{Appendix}
\label{sec:appendix} 

\subsection{Input Formats in Related Work}
\label{sec:related_work_formats}

\paragraph{\squadprompt (T5's prompt for \textsc{SQuAD})}

\begin{compactitem}
\item \texttt{question:} [\textit{question}] \texttt{context:} [\textit{paragraph}]
\end{compactitem}

\noindent\textbf{\unifiedqa's prompts} (basis for \unifiedqaprompt)

\begin{compactitem}
\item \textbf{Multiple-choice QA:} [\textit{question}] \texttt{\textbackslash{}\textbackslash{}n (A)} [\textit{choice}$_1$] \texttt{(B)} [\textit{choice}$_2$]...
\item \textbf{Extractive QA:} [\emph{question}] \texttt{\textbackslash{}\textbackslash{}n} [\emph{paragraph}]
\end{compactitem}

\paragraph{\textsc{UniFew}}

\begin{compactitem}
\item \textbf{Single text classification:} \texttt{Topic? \textbackslash{}\textbackslash{}n (A)} [\textit{class$_1$}] \texttt{(B)}  [\textit{class$_2$}] \texttt{(C)} [\textit{class$_3$}] \textbackslash{}\textbackslash{}n [\textit{document}]
\item \textbf{Sentence-pair classification:} [\textit{sentence$_1$}] \texttt{Is} [\textit{sentence$_1$}] \texttt{? \textbackslash{}\textbackslash{}n  (A) Yes (B) No (C) Maybe}
\item \textbf{Relation classification:}
[\textit{mention$_1$}] \texttt{to} [\textit{mention$_1$}] \texttt{? \textbackslash{}\textbackslash{}n  (A)} [\textit{class$_1$}] \texttt{(B)}  [\textit{class$_2$}] \texttt{(C)} [\textit{class$_3$}] 
\item \textbf{Entity recognition:} \texttt{What is the type of the entity between the \# marks? \textbackslash{}\textbackslash{}n} [\textit{text...} \texttt{\#}\textit{mention$_1$}\texttt{\#} \textit{...text...} 
\texttt{\#}\textit{mention$_2$}\texttt{\#} \textit{...text}]
\end{compactitem}

\subsection{Prompts Used in \sect{sec:scaling}}
\label{sec:finall_prompts}

We use \simtfive for \esnli, and \unifiedqaprompt for other datasets:
\begin{itemize}
    \item \textbf{\esnli} \texttt{explain nli hypothesis:} [\textit{hypothesis}] \texttt{premise:} [\textit{premise}]
    \item \textbf{\ecqa:} \texttt{explain} [\textit{question}] \texttt{\textbackslash{}\textbackslash{}n (A)} [\textit{choice}$_1$] \texttt{(B)} [\textit{choice}$_2$] \texttt{(C)} [\textit{choice}$_3$] \texttt{(D)} [\textit{choice}$_4$] \texttt{(E)} [\textit{choice}$_5$] 
    \item \textbf{\comve:} \texttt{explain what is more nonsensical? \textbackslash{}\textbackslash{}n choice1:} [\textit{sentence$_1$}] \texttt{choice2:} [\textit{sentence$_2$}] 
    \item \textbf{\sbic:} \texttt{explain what is this? \textbackslash{}\textbackslash{}n post:} [\textit{post}]
\end{itemize}

Prompts for \textsc{GPT-3} are given in Figure \ref{fig:gpt3_prompts}. The output always takes the form of ``[\emph{answer/label}] \texttt{because} [\emph{explanation}]''.

\begin{figure}[t]
\small
        \begin{tcolorbox}[width=\columnwidth, colframe=gray!20, colback=gray!10, halign=left, boxsep=0pt, segmentation style=black]

        \textbf{\textsc{SBIC}:}
        
        \vspace{2mm}
        
        \texttt{Is the following post offensive? and why?}
    
        \texttt{Post:} [\textit{post}]
            
        \texttt{Answer:} [\textit{Yes} | \textit{No}]
            
        \texttt{Reason:} [\textit{explanation}]
    
        \texttt{}
        
        \tcbline
        
        \texttt{}

        \textbf{\textsc{E-SNLI}:}
        
        \vspace{2mm}

       \texttt{Answer the question and provide a reason why the answer is correct.}
    
        [\textit{premise}]
        
        \texttt{Question: Is} [\textit{hypothesis}]?
            
        \texttt{Answer:} [\textit{Yes} | \textit{No} | \textit{Maybe}]
            
        \texttt{Reason:} [\textit{explanation}]

        \texttt{}
        
        \tcbline
        
        \texttt{}

        \textbf{\textsc{ECQA}:}
        
        \vspace{2mm}

        \texttt{Answer the question from the provided choices, and provide a reason why the answer is correct.}
    
        \texttt{Question:} [\textit{question}]
        
        \texttt{Choices:} [\textit{choices}]
            
        \texttt{Answer: } [\textit{one of the choices}]
            
        \texttt{Reason:} [\textit{explanation}]

        \texttt{}
        
        \tcbline
        
        \texttt{}

         \textbf{\textsc{ComVE}:}
         
        \vspace{2mm}

        \texttt{Which of the two choices makes more sense? and why?}
    
        \texttt{Choice1:} [\textit{choice1}]
        
        \texttt{Choice2:} [\textit{choice2}]
            
        \texttt{Answer:} [\textit{Choice1} | \textit{Choice2}]
            
        \texttt{Reason:} [\textit{explanation}]
        
        \end{tcolorbox}
    \caption{GPT-3 prompt templates for all datasets.}
\label{fig:gpt3_prompts}
\end{figure}

\newpage

\begin{table}[!h]
 \resizebox{\columnwidth}{!}{  
\begin{tabular}{l|lp{5cm}}
\toprule
\textbf{\benchmark Task} & \multicolumn{2}{l}{\textbf{Similar T5 Pretraining Tasks}} \\
\midrule
\makecell[tl]{\textsc{E-SNLI}} & \makecell[tl]{\textsc{MNLI}\\  \cite{williams-etal-2018-broad}}  & Classify the entailment relation between two sequences                    \\
\arrayrulecolor{black!20}\midrule
\makecell[tl]{\textsc{ECQA}} & \makecell[tl]{\textsc{RECORD}\\ \cite{Zhang2018ReCoRDBT}} & Answer a cloze-style query about a passage given entities in it         \\
\arrayrulecolor{black!20}\midrule
\makecell[tl]{\textsc{ComVE}} & \makecell[tl]{\textsc{COPA}\\ \cite{Roemmele2011COPA}}  & Select one of two sequences as the cause/effect of a premise \\
\arrayrulecolor{black!20}\midrule
\makecell[tl]{\textsc{SBIC}} & \makecell[tl]{\textsc{COLA}\\  \cite{warstadt-etal-2019-neural}}  & Classify a sentence as acceptable or not       \\
\arrayrulecolor{black}\bottomrule
\end{tabular}
}
\caption{The first column shows tasks that we have included in \benchmark. Tasks on the right are included in T5's pretraining and they are similar to \benchmark's tasks. We explore self-rationalization prompts for \benchmark's tasks based on the tasks on the right, and compare them to prompts designed as span infilling and QA  (\sect{sec:study_various_prompts}). 
}
\label{tab:similar_tasks}
\end{table}
\begin{table}[t]
    \centering
    \small
    \vspace{-3.65cm}
    \resizebox{\columnwidth}{!}{ 
    \begin{tabular}{cc}
      \toprule
      \textbf{GPUs} & 8 NVIDIA A100s 48 GB on Google Cloud\\
      \midrule
      \textbf{Implementation} & \url{https://github.com/allenai/feb} \\
      \bottomrule
    \end{tabular}}
    \vspace{1mm}
    \resizebox{\columnwidth}{!}{ 
    \begin{tabular}{ll}
        \toprule
        \textbf{Hyperparameter} & \textbf{Assignment}  \\
        \midrule
        max step number & 300 \\
        \midrule
        batch size & 4 (1 for T5/\unifiedqa-3B) \\
        \midrule 
        grad.\ accumulation steps & 1 (4 for T5/\unifiedqa-3B)\\
        \midrule 
        learning rate & 3e-5\\
          \midrule
        learning rate scheduler & linear \\
        \midrule 
        warmup steps & 0 \\
        \midrule
        decoding & greedy\\
        \bottomrule
    \end{tabular}
    }
    \caption{Hyperparameters used in our experiments.} 
    \label{tab:hyperparameters}
\end{table}

\begin{table}[!h]
\centering
\vspace{-2.5in}
\resizebox{\columnwidth}{!}{  
\begin{tabular}{llrr}
\toprule
            &  & \textbf{Accuracy} & \textbf{BERTscore} \\
\midrule
\multirow{4}{*}{\rotatebox{90}{\textsc{CoS-E}}} &\infilling (b) & 34.3$_{0.4}$ & 29.6$_{0.3}$ \\
&\infilling (n) & 40.1$_{0.4}$ & 34.7$_{0.3}$ \\
&\simtfive      & 51.7$_{0.4}$ & 44.6$_{0.4}$ \\
& \squadprompt        & 51.1$_{0.3}$ & 44.1$_{0.3}$ \\
&\unifiedqaprompt  & \textbf{60.0}$_{0.3}$ & \textbf{48.6}$_{0.3}$ \\
\arrayrulecolor{black}\bottomrule
\end{tabular}}
\caption{A comparison of all prompt types introduced in \sect{sec:study_various_prompts} on \cose. We do not support using \cose in the future given the reported issues with it \cite{narang2020wt5, wiegreffe-marasovic-2021-review}, especially since \ecqa is introduced.}
\label{tab:appendix_cos_various}
\end{table}

\begin{table}[!ht]
\centering
\vspace{-2.5in}
 \resizebox{\columnwidth}{!}{  
\begin{tabular}{llrr}
\toprule
& \textbf{Size} & \textbf{Accuracy}  & \textbf{BERTscore} \\
\midrule
\multirow{5}{*}{\rotatebox{90}{\textsc{CoS-E}}} & \textsc{\small Base}  & 58.3$_{0.3}$ &	50.4$_{0.2}$  \\
                      &\textsc{\small Large} & 69.4$_{0.3}$ &	60.1$_{0.3}$   \\
                      &\textsc{\small 3B} & \textbf{75.4}$_{0.3}$ & \textbf{65.3}$_{0.3}$  \\
\arrayrulecolor{black!20}\cmidrule{2-4}
                      &\textsc{\small GPT-3} & 68.4$_{1.3}$ & 59.5$_{1.2}$  \\ 
\arrayrulecolor{black}\bottomrule
\end{tabular}}
\caption{The effect of scaling the \unifiedqa model size on self-rationalization of \cose. We do not support using \cose in the future given the reported issues with it \cite{narang2020wt5, wiegreffe-marasovic-2021-review}, especially since \ecqa is introduced.}
\label{tab:appendix_cose_scale}
\end{table}
\begin{table}[t]
\centering
\resizebox{\columnwidth}{!}{  
\begin{tabular}{llrr}
\toprule
                  & \textbf{Prompt}                                        & \textbf{Accuracy}  & \textbf{BERTscore}\\
\midrule
\multirow{4}{*}{\rotatebox{90}{\textsc{E-SNLI}}}  & Is...?      & 38.7$_{0.4}$         & 34.7$_{0.4}$              \\
                        & + tags   & 48.2$_{0.6}$          & 43.2$_{0.6}$              \\
                        \arrayrulecolor{black!20}\cmidrule{2-4}
                        & What is...? & 60.7$_{0.8}$          & 54.7$_{0.8}$              \\
                        & + tags   & \textbf{77.9}$_{0.3}$          & \textbf{70.1}$_{0.3}$              \\
\arrayrulecolor{black}\midrule
\multirow{2}{*}{\rotatebox{90}{\ecqa}} & \textsc{\small \textsc{SQuAD}$_{\textnormal{\scriptsize T5}}$}  & \textbf{36.5}$_{0.3}$ & \textbf{32.4}$_{0.3}$ \\
                        \arrayrulecolor{black!20}\cmidrule{2-4}
                        & \textsc{\small Random Baseline} & \makecell[c]{20.0} & \makecell[c]{-}\\
\arrayrulecolor{black}\midrule
\multirow{4}{*}{\rotatebox{90}{\textsc{ComVE}}}   & Is...? & 50.4$_{0.2}$ &	45.5$_{0.1}$ \\
& + tags     & 50.2$_{0.1}$           & 45.3$_{0.1}$              \\
                        \arrayrulecolor{black!20}\cmidrule{2-4}
                        & What is...? & 50.5$_{0.2}$ &	45.7$_{0.2}$ \\
                        & + tags      & \textbf{54.5}$_{0.5}$ & \textbf{49.2}$_{0.4}$ \\
 \arrayrulecolor{black}\midrule
\multirow{4}{*}{\rotatebox{90}{\textsc{SBIC}}}   & Is...?      & 63.4$_{0.6}$           & 61.1$_{0.6}$              \\
                        & + tags   & 63.8$_{0.5}$ &	61.7$_{0.5}$\\
                        \arrayrulecolor{black!20}\cmidrule{2-4}
                        & What is...?    & 66.7$_{0.5}$           & 64.3$_{0.5}$              \\
                        & + tags   & \textbf{67.0}$_{0.5}$	& \textbf{64.6}$_{0.6}$ \\
 \arrayrulecolor{black}\bottomrule
\end{tabular}
}
\caption{A comparison between \squadprompt prompts with ``Is...?'' and ``What is...?'' questions. See \sect{sec:qa_prompts} for more info. We also inspect the effects of adding answer choices and \emph{tags} to the input. Tags are a single word descriptions of the input elements; e.g., \textsc{\small E-SNLI}'s tags are ``premise:'' / ``hypothesis:'' before premise / hypothesis.}
\label{tab:squad_qa_prompts}
\vspace{128in}
\end{table}
\begin{table*}[!ht]
 \resizebox{\textwidth}{!}{  
 \centering
\begin{tabular}{p{1.125\textwidth}}
\toprule
\textbf{Sentence1:} The stove was cleaned with a cleaner. \textbf{Sentence2:} The stove was cleaned with a mop. \\
\textbf{Nonsensical Sentence}: Sentence2 \textbf{Explanation:} A mop is too large to clean the stove.\\
\midrule
\midrule
\textbf{Prompt:} \infilling $\times$ \textsc{Basic} \\
\textbf{Input:} \texttt{explain sensemaking choice1:} \emph{The stove was cleaned with a cleaner.} \texttt{choice2:} \emph{The stove was cleaned with a mop.} \texttt{\textless{}extra\_id\_0\textgreater{} because \textless{}extra\_id\_1\textgreater{}} \\
\textbf{Output:} \texttt{\textless{}extra\_id\_0\textgreater{}} \emph{choice2} \texttt{\textless{}extra\_id\_1\textgreater{}} \emph{A mop is too large to clean the stove.} \texttt{\textless{}extra\_id\_2\textgreater{}} \\
\arrayrulecolor{black!20}\midrule
\textbf{Prompt:} \infilling $\times$ \textsc{Natural Sounding} \\
\textbf{Input:} \texttt{explain sensemaking choice1:} \emph{The stove was cleaned with a cleaner.} \texttt{choice2:} \emph{The stove was cleaned with a mop.} \texttt{It is \textless{}extra\_id\_0\textgreater{} that choice2 is less common because \textless{}extra\_id\_1\textgreater{}} \\ 
\textbf{Output:} \texttt{\textless{}extra\_id\_0\textgreater{}} \emph{True} \texttt{\textless{}extra\_id\_1\textgreater{}} \emph{A mop is too large to clean the stove.} \texttt{\textless{}extra\_id\_2\textgreater{}}      \\
\arrayrulecolor{black}\midrule
\midrule
\textbf{Prompt:} \simtfive $\times$ \textsc{COPA} \\
\textbf{Input:} \texttt{explain sensemaking choice1:} \emph{The stove was cleaned with a cleaner.} \texttt{choice2:} \emph{The stove was cleaned with a mop.} \texttt{Less common is choice2}\\
\textbf{Output:} \emph{True} \texttt{because} \emph{a mop is too large to clean the stove.} \\      
\midrule
\midrule
\textbf{Prompt:}  \squadprompt $\times$ \textsc{Yes/No} + \textsc{Tags}\\
\textbf{Input:} \texttt{explain sensemaking question:  Is choice2 more nonsensical? context: choice1:} \emph{The stove was cleaned with a cleaner.} \texttt{choice2:} \emph{The stove was cleaned with a mop.} \\         
\textbf{Output:} \emph{Yes} \texttt{because} \emph{a mop is too large to clean the stove.} \\  
\arrayrulecolor{black!20}\midrule
\textbf{Prompt:} \squadprompt $\times$ \textsc{What is...?} + \textsc{Tags} \\
\textbf{Input:} \texttt{explain sensemaking question: What is more nonsensical? context: choice1:} \emph{The stove was cleaned with a cleaner.} \texttt{choice2:} \emph{The stove was cleaned with a mop.} \\
\textbf{Output:} \emph{choice2} \texttt{because} \emph{a mop is too large to clean the stove.} \\
\arrayrulecolor{black}\midrule
\midrule
\textbf{Prompt:} \unifiedqaprompt $\times$ \textsc{Yes/No} \\
\textbf{Input:} \texttt{explain is choice2 more nonsensical? \textbackslash{}\textbackslash{}n} \emph{The stove was cleaned with a cleaner.} \emph{The stove was cleaned with a mop.}\texttt{</s>} \\
\textbf{Output:}  \emph{yes} \texttt{because} \emph{a mop is too large to clean the stove.}    \\
\arrayrulecolor{black!20}\midrule
\textbf{Prompt:} \unifiedqaprompt $\times$ \textsc{Yes/No} + \textsc{Tags} \\
\textbf{Input:} \texttt{explain is choice2 more nonsensical? \textbackslash{}\textbackslash{}n choice1:} \emph{The stove was cleaned with a cleaner.} \texttt{choice2:} \emph{The stove was cleaned with a mop.}\texttt{</s>} \\
\textbf{Output:}  \emph{yes} \texttt{because} \emph{a mop is too large to clean the stove.}    \\
\arrayrulecolor{black!20}\midrule
\textbf{Prompt:} \unifiedqaprompt $\times$ \textsc{Yes/No} + \textsc{Tags} + \textsc{Choices} \\
\textbf{Input:} \texttt{explain is choice2 more nonsensical? \textbackslash{}\textbackslash{}n (A) yes (B) no \textbackslash{}\textbackslash{}n choice1: } \emph{The stove was cleaned with a cleaner.} \texttt{choice2:} \emph{The stove was cleaned with a mop.}\texttt{</s>}\\
\textbf{Output:} \emph{yes} \texttt{because} \emph{a mop is too large to clean the stove.}   \\               
\arrayrulecolor{black!20}\midrule
\textbf{Prompt:} \unifiedqaprompt $\times$ \textsc{What is...?} \\
\textbf{Input:} \texttt{explain what is more nonsensical? \textbackslash{}\textbackslash{}n} \emph{The stove was cleaned with a cleaner.} \emph{The stove was cleaned with a mop.}\texttt{</s>} \\
\textbf{Output:} \emph{choice2} \texttt{because} \emph{a mop is too large to clean the stove.} \\
\arrayrulecolor{black!20}\midrule
\textbf{Prompt:} \unifiedqaprompt $\times$ \textsc{What is...?} + \textsc{Tags}\\
\textbf{Input:} \texttt{explain what is more nonsensical? \textbackslash{}\textbackslash{}n choice1:} \emph{The stove was cleaned with a cleaner.} \texttt{choice2:} \emph{The stove was cleaned with a mop.}\texttt{</s>} \\
\textbf{Output:} \emph{choice2} \texttt{because} \emph{a mop is too large to clean the stove.} \\
\arrayrulecolor{black!20}\midrule
\textbf{Prompt:} \unifiedqaprompt $\times$ \textsc{What is...?} + \textsc{Tags} + \textsc{Choices} \\
\textbf{Input:} \texttt{explain what is more nonsensical? \textbackslash{}\textbackslash{}n (A) choice1 (B) choice2 \textbackslash{}\textbackslash{}n choice1:} \emph{The stove was cleaned with a cleaner.} \texttt{choice2:} \emph{The stove was cleaned with a mop.}\texttt{</s>}\\
\textbf{Output:} \emph{choice2} \texttt{because} \emph{a mop is too large to clean the stove.} \\
\arrayrulecolor{black}\bottomrule
\end{tabular}
}
\caption{\comve self-rationalization prompts that we design and test. \infilling marks span-filling prompts; \simtfive prompts made by following the most similar T5 pretraining task (Table \ref{tab:tasks}); \squadprompt prompts designed following \textsc{\small SQuAD}'s formatting in T5 pretraining; and   \unifiedqaprompt prompts made following \textsc{\small UnifiedQA}. This table shows variations of these prompt types. We refer to spans ``choice1:''/``choice2:'' as \textsc{\small Tags}, and to ``(A) yes (B) no''/``(A) choice1 (B) choice2'' as \textsc{\small Choices}. \textsc{\small Yes/No} and \textsc{\small What is...?} refer to a question type. Following \citet{Hendrycks2021MeasuringMM}, we add </s> to the end of our \unifiedqaprompt prompts. More info in \sect{sec:study_various_prompts}.} 
\label{tab:prompts_comve}
\end{table*}

\end{document}